\documentclass[11pt]{article}
\usepackage[table]{xcolor}
\usepackage[final]{acl}

\usepackage{times}
\usepackage{latexsym}

\usepackage[T1]{fontenc}

\usepackage[utf8]{inputenc}

\usepackage{microtype}

\usepackage{inconsolata}

\usepackage{graphicx}
\usepackage{amsmath}
\usepackage{latexsym}
\usepackage{tabularx} 
\usepackage{booktabs}
\usepackage{multirow}
\usepackage{subfigure}
\usepackage{amsmath}
\usepackage{amssymb}
\usepackage{array}
\usepackage{bm}
\usepackage{xparse}
\usepackage{mathtools}
\usepackage{pifont}
\usepackage{enumitem}
\usepackage{tcolorbox}
\tcbuselibrary{breakable}
\usepackage{listings}

\usepackage{arydshln}
\usepackage{hyperref}
\usepackage{algorithm}

\usepackage{colortbl}
\usepackage{booktabs}
\usepackage{siunitx}

\usepackage{algpseudocode}
\usepackage{xcolor}
\usepackage{amsmath}
\usepackage{amssymb}
\usepackage[italicComments=true, indLines=true]{algpseudocodex} 
\usepackage{xcolor}
\definecolor{darkred}{RGB}{160, 30, 30}
\definecolor{graycomment}{RGB}{120, 120, 120}
\usepackage{inputenc}
\newcommand{\gcell}[1]{\cellcolor[HTML]{F1F8E9}#1}
\algrenewcommand{\algorithmiccomment}[1]{\hfill \textit{\textcolor{graycomment}{// #1}}}
\title{Pushing the Limits of LLM Tool Calling via Experiential Knowledge Integration and Activation}

\author{
 \textbf{Yupu Hao\textsuperscript{1,2}},
 \textbf{Zhuoran Jin\textsuperscript{1,2}},
 \textbf{Huanxuan Liao\textsuperscript{1,2}},
 \textbf{Kang Liu\textsuperscript{1,2}},
 \textbf{Jun Zhao\textsuperscript{1,2}\thanks{Corresponding Author}}
\\
 \textsuperscript{1}The Key Laboratory of Cognition and Decision Intelligence for Complex Systems, \\Institute of Automation, Chinese Academy of Sciences, Beijing, China \\
 \textsuperscript{2}School of Artificial Intelligence, University of Chinese Academy of Sciences, Beijing, China
\\
\{haoyupu2023, liaohuanxuan2023\}@ia.ac.cn, \{zhuoran.jin, kliu, jzhao\}@nlpr.ia.ac.cn
}

\begin{document}
\maketitle
\begin{abstract}
Large language models (LLMs) rely on tool use to act as autonomous agents, yet often fail in multi-step execution due to insufficient tool-related knowledge and ineffective knowledge activation. Therefore, we present a systematic study on how knowledge influences tool-use performance, covering the stages of \textbf{knowledge acquisition, activation, and internalization}. In the knowledge acquisition stage, we acquire and evaluate various forms of experiential knowledge, and our analysis shows that simple instance-level knowledge can already provide strong and reliable gains, while abstract intent-level knowledge offers limited benefits. At inference time, to activate knowledge, we find that prompting LLM to expand the depth of reasoning yields diminishing returns, whereas expanding the width of reasoning by parallel sampling with aggregation more effectively activates latent experiential knowledge. At training time, for knowledge internalization, post-training with knowledge-augmented data further improves performance, with reinforcement learning outperforming supervised fine-tuning. Based on these insights, we propose the \textbf{K}nowledge-\textbf{A}ugmented \textbf{T}ool \textbf{E}xecution (\textbf{KATE}), a knowledge-augmented tool execution framework that integrates experiential knowledge with reasoning-width-expanded inference and knowledge-aware training. Experiments on BFCL-V3 and AppWorld demonstrate consistent and substantial improvements over strong baselines across model scales. Our Code is available at \url{https://github.com/hypasd-art/KATE}.

\end{abstract}

\section{Introduction}
Tool use has emerged as a cornerstone capability for transforming large language models (LLMs) into practical intelligent agents \cite{DBLP:conf/emnlp/LiZ000YLHL23, DBLP:journals/tmlr/MialonDLNPRRSDC23, liagentic}. LLMs increasingly rely on tool calling to execute actions, access external information \cite{DBLP:journals/corr/abs-2503-09516}, and serve as autonomous agents \cite{DBLP:journals/corr/abs-2503-23037}. However, existing approaches largely treat tool use as a problem of prompt design \cite{DBLP:conf/nips/ShinnCGNY23}, API documents specification \cite{DBLP:conf/iclr/QuDWCWY0W25}, or supervised or unsupervised alignment \cite{DBLP:conf/iclr/Liu0ZHYL0GLY0WN25, DBLP:journals/corr/abs-2503-23383, DBLP:journals/corr/abs-2505-00024, DBLP:journals/corr/abs-2510-10197}, implicitly assuming that models already possess sufficient experiential knowledge for tool execution. In practice, however, failures in tool use often stem not from reasoning incapability alone, but from the lack of concrete, executable experience, such as parameter constraints, scenario-specific operation patterns, and error recovery strategies. 

While prior work has explored knowledge augmentation for general reasoning \cite{DBLP:conf/emnlp/WangYXQD00GJX0C24}, the role of \emph{experiential knowledge} \cite{DBLP:journals/corr/abs-2508-06433, DBLP:journals/corr/abs-2508-16153} in tool execution remains largely underexplored. In particular, it is unclear (i) which forms of knowledge are most effective for tool use, (ii) how knowledge within the system should be activated during inference, and (iii) whether there are additional gains if the knowledge is internalized into model parameters through training. Addressing these questions requires a systematic investigation that spans retrieval, inference-time reasoning, and training-time optimization, which is an aspect missing from existing studies.

To bridge this gap, we conduct the first systematic study of experiential knowledge in tool execution, examining how different types of experiential knowledge can be acquired, activated, and internalized within large language models. Unlike prior works that primarily focuses on designing specific knowledge representations construction process and more fine-grained retrieval mechanisms \cite{cao2025remembermerefineme, DBLP:journals/corr/abs-2508-06433, DBLP:conf/icml/WangMFN25}, we emphasizes a unified and principled understanding of how knowledge functions throughout the entire pipeline. We organize our investigation along two complementary dimensions: \textbf{Knowledge Acquisition and Integration} and \textbf{Knowledge Activation and Utilization}. From the perspective of \textbf{knowledge acquisition}, we extract and categorize four types of experiential knowledge which includes \emph{instance-level} Scenario Trajectory Knowledge and Experience Summary Knowledge, as well as \emph{intent-level} Script-Style Intent Clustering Knowledge and Textual-Style Intent Clustering Knowledge, and design a unified retrieval mechanism to integrate them at inference time. Through extensive experiments, we demonstrate that instance-level knowledge consistently yields the largest performance gains, indicating that concrete execution traces or its corresponding description provide more actionable guidance than abstract intent descriptions for tool-using agents. The results demonstrate that \emph{high-quality execution trajectories alone are sufficient to yield substantial performance improvements in tool use}. From the perspective of \textbf{knowledge activation}, we investigate how to effectively elicit and utilize such knowledge. At inference time, we compare depth-based hint prompting with width-based parallel sampling, \emph{revealing a clear advantage of expanding the reasoning width over increasing depth of reasoning}. While explicit prompts engineering provide diminishing returns as model capability scales, parallel sampling with aggregation substantially improves tool-calling accuracy, suggesting that much of the model’s experiential knowledge remains latent under deterministic decoding. At training time, we further show that \emph{fine-tuning with knowledge-augmented data enables deeper internalization of experiential knowledge, yielding additional gains beyond context-based retrieval alone}. We adopt both supervised fine-tuning (SFT) and reinforcement learning (RL), and find that RL leads to more substantial performance improvements.

Based on these findings, we propose \textbf{KATE} (\textbf{K}nowledge-\textbf{A}ugmented \textbf{T}ool \textbf{E}xecution), a unified framework that systematically incorporates experiential knowledge across acquisition, activation, and training stages. KATE integrates instance-level knowledge with width-based parallel sampling to effectively activate latent knowledge during inference, and further internalizes such knowledge through post-training. Empirical results demonstrate that KATE achieves significant and consistent improvements in tool-use accuracy across model scales and task settings.

Our work makes three key contributions:
\begin{itemize}
    \item We systematically investigate how different granularities of tool-usage knowledge affect tool execution. By designing multiple experiential knowledge acquisition strategies, we show that simple, high-quality instance-level knowledge alone can already provide effective improvements.

    \item We study how tool-related knowledge is activated during both inference and training. We analyze reasoning depth and width and find that parallel sampling with aggregation more effectively activates latent knowledge. And post-training yields additional gains beyond context-based knowledge injection.

    \item Our method \textbf{KATE} is a unified knowledge-augmented tool execution framework that integrates instance-level experience with width-expanded inference and knowledge-aware training. KATE achieves state-of-the-art performance in both training-free and training-based settings. On the Qwen3-8B model of dataset BFCL-V3, our method improves average performance by 15\% compared to direct tool use.
\end{itemize}

\section{Preliminary}
Multi-turn tool-utilization by LLMs can be formulated as a Markov decision process (MDP). At interaction step $t$, conditioned on the set of available tools $\mathcal{T}$, the system prompt $S$ and the previous dialogue history $\mathcal{H}_t$, the core objective of the LLM $P$ is to predict the next action $o_{t+1}$ based on the current context:

\begin{equation}
\footnotesize
    o_{t+1} = P(\mathcal{T}, S, \mathcal{H}_t)
\end{equation}

where $o_{t+1}$ representing either a tool invocation $c_{t+1}$ or a final natural language response $a_{t+1}$.
 
After the model emits $o_{t+1}$, the environment returns external feedback $r_{t+1}$, which can be categorized as either a \textbf{tool execution response} $r^{\text{env}}_{t+1}$ or a \textbf{user reply} $r^{\text{user}}_{t+1}$. The dialogue history is then updated as follows:

\begin{equation}
\footnotesize
    \mathcal{H}_{t+1} = \mathcal{H}_t \cup o_{t+1} \cup r_{t+1}
\end{equation}

This updated state serves as the context for the subsequent decision step, thereby completing the Markovian interaction loop.

\section{Method}
\label{sec:analysis_setting}
We present the study and method of experiential knowledge in tool execution, examining how it is acquired, activated, and internalized.

\subsection{Knowledge Acquisition and Integration}
Knowledge plays an essential role in successful tool execution. We systematically investigate how different types of experiential knowledge influence model performance, as well as how such knowledge can be efficiently retrieved and utilized during inference through a structured knowledge base. 
\subsubsection{Knowledge Base Construction} 
To study the role of different experiential knowledge, we categorize experiential knowledge into two levels based on granularity: \textbf{Instance-level Knowledge}, which provides concrete, example-specific guidance, and \textbf{Intent-level Knowledge}, which captures higher-level abstractions of task objectives and decision patterns.

For Instance-level Knowledge, we consider two forms: (1) Scenario Trajectory Knowledge \textbf{(ST)}: Ground-truth tool execution trajectories are directly used as knowledge inputs during inference, providing explicit step-by-step guidance. (2) Experience Summary Knowledge \textbf{(ES)}: An LLM is prompted with paired user queries and ground-truth trajectories from the training data to generate concise, high-level operational guidelines in textual form.

For intent-level knowledge, we observe that each user query in a scenario naturally reflects a specific intent (e.g., information retrieval, shopping online). These intents serve as the fundamental components of more complex goals, and tool invocation patterns are often consistent within the same intent category.
Thus, we construct two forms of intent-level knowledge: (1) Script-Style Intent Clustering Knowledge \textbf{(SIC)}: We generate the intents of user questions, cluster training examples accordingly and summarize tool-usage scripts with an LLM in a semi-structured form. (2) Textual-Style Intent Clustering Knowledge \textbf{(TIC)}: We additionally provide unstructured, natural-language descriptions that capture the operational strategies for each intent category based on the cluster result.  The details of knowledge base construction are in Appendix~\ref{sec:appendix_knowledge_construction} and the examples of user's questions with retrieval knowledge are in Appendix~\ref{sec:appendix_example}.

To construct knowledge base $\mathcal{K}$, we build the retrieval base by encoding and storing the user queries into vector representations using a language model encoder for Instance-level Knowledge. For Intent-level knowledge, we encode the inferred user intents $I$ rather than the raw queries. Together, these knowledge forms differ in both granularity and representation, enabling a systematic study of how experiential knowledge influences tool-use learning and inference.

\subsubsection{Knowledge Retrieval}
During inference, if the feedback $r_{t+1}$ is a user query $r^{\text{user}}_{t+1}$, we automatically retrieve relevant knowledge from an external knowledge base. 
\begin{figure*}[tp]
    \centering
    \includegraphics[width=1\textwidth]{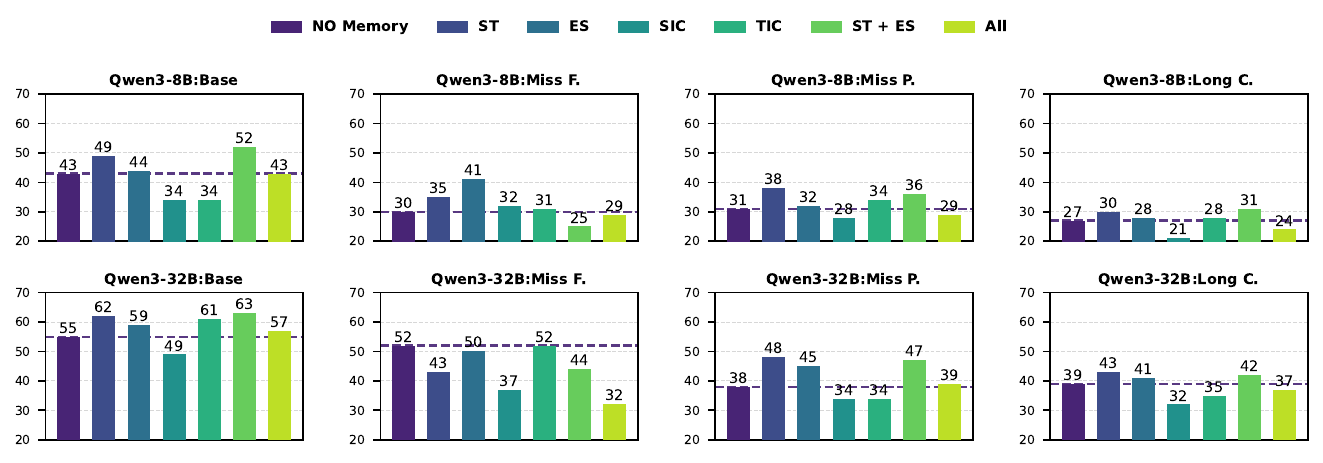}
    \caption{The augmentation results of different experiential knowledge. ``All'' indicates incorporating all the experiential knowledge.}
    \label{fig:knowledge_augment_result}
\end{figure*}
For Instance-level Knowledge, we employ the same language model encoder to map the user query $r^{\text{user}}_{t+1}$ into a vector representation and perform similarity matching against the stored knowledge embeddings. Knowledge entries whose similarity scores exceed a predefined threshold $p$ are ranked, and the top-$K$ entries are selected as retrieved knowledge. These retrieved entries are then concatenated with the original user query and provided to the model as augmented input.

For intent-level knowledge, we first prompt the model to explicitly infer the user’s current intent $I_{t+1}$. The inferred intent is subsequently encoded and used as the query to retrieve intent-level knowledge. The knowledge entry corresponding to the most similar intent is selected as the final retrieval result ($K$=1). 

Formally, the retrieval operation is defined as:
\begin{equation}
\footnotesize
\mathcal{R}(Q) = \operatorname{Top\text{-}K}\Big(\mathbf{k}_j \big| \mathbf{k}_j \in \mathcal{K}, \text{sim}(Q, \mathbf{k}_j) \ge p \Big)
\end{equation}
where $Q$ denotes either the user query $r^{\text{user}}_{t+1}$ or the inferred intent $I_{t+1}$, and $\mathcal{K}$ represents the knowledge base. For intent-level retrieval, we set $K=1$.

When a user message is observed, the retrieved knowledge is incorporated into the interaction as:
\begin{equation}
\footnotesize
r^{\text{re}}_{t+1} = r_{t+1} \cup \mathcal{R}(Q), \quad \text{if } r_{t+1} = r^{\text{user}}_t
\end{equation}

and the dialogue history augmented with retrieved knowledge is updated as:
\begin{equation}
\footnotesize
\mathcal{H}^{\text{re}}_{t+1} = \mathcal{H}^{\text{re}}_t \cup {o_{t+1}, r^{\text{re}}_{t+1}}
\end{equation}

We conduct analysis experiments on the BFCL-V3 \cite{DBLP:conf/icml/PatilMYJSSG25} benchmark. We evaluate our approach on Qwen3-8B and Qwen3-32B \cite{DBLP:journals/corr/abs-2505-09388}, systematically comparing different experiential knowledge types and integration settings, as shown in Figure~\ref{fig:knowledge_augment_result}. The experimental results show that:
(1) \textbf{Instance-level knowledge consistently yields greater performance improvements than intent-level knowledge}. This is likely because trajectory-level information provides fine-grained and directly executable guidance, whereas intent-level knowledge requires multi-step abstraction and intent matching, which may introduce additional errors due to imperfect intent inference by LLMs. 
(2) \textbf{Scenario Trajectory Knowledge (ST) and Experience Summary Knowledge (ES), as well as their combination, exhibit comparable overall performance}, with their relative effectiveness varying across different tasks and model backbones. This suggests that no single form of instance-level knowledge universally dominates, and that task-specific and model-specific adaptation is necessary to achieve optimal performance. 
(3) \textbf{We observe that simply stacking multiple types of knowledge does not guarantee further gains}. Naive combinations may lead to redundancy or interference among knowledge sources, underscoring that effective integration and utilization strategies are more important than the quantity of experiential knowledge provided. This observation motivates the need for structured retrieval and selective activation mechanisms, rather than indiscriminate knowledge aggregation.

\subsection{Knowledge Activation and Utilization}

Given a fixed amount of knowledge, a central question is how to more effectively activate a model’s tool-use capabilities to produce reliable and accurate tool outputs. We investigate this problem from both the training-time and inference-time perspectives.

\subsubsection{Inference-Time}
 
At inference time, we identify the \textit{reasoning depth} and \textit{reasoning width} as two key factors that influence knowledge activation. 
Based on this observation, we explore two main strategies: \textbf{Depth-based Prompt-Hint Activation}, which encourages deeper and more detailed reasoning, and \textbf{Width-Based Parallel Sampling with Aggregation}, which expands the reasoning space by exploring multiple candidate trajectories.

\textbf{Depth-based Prompt-Hint Activation.} Prompt engineering enhances model reasoning by shaping the input prompt. depth-based Prompt-Hint methods aim to increase reasoning depth by explicitly providing guidance that encourages structured reasoning patterns. Concretely, after each tool execution, a hint is appended as a user-role message before the next tool decision, prompting the model to explicitly consider tool selection and action planning. Based on the error-prone scenarios identified in the analysis in \citeauthor{DBLP:conf/icml/PatilMYJSSG25}\shortcite{DBLP:conf/icml/PatilMYJSSG25}, we design the three prompt hints to target the model’s common failure modes. These hints are constructed from three complementary perspectives: \emph{intent}, \emph{reflection}, and \emph{state}. Specifically, the model is instructed to reason over these aspects and base subsequent tool calls on the resulting structured analysis. 

\begin{figure*}[tp]
    \centering
    \includegraphics[width=1\textwidth]{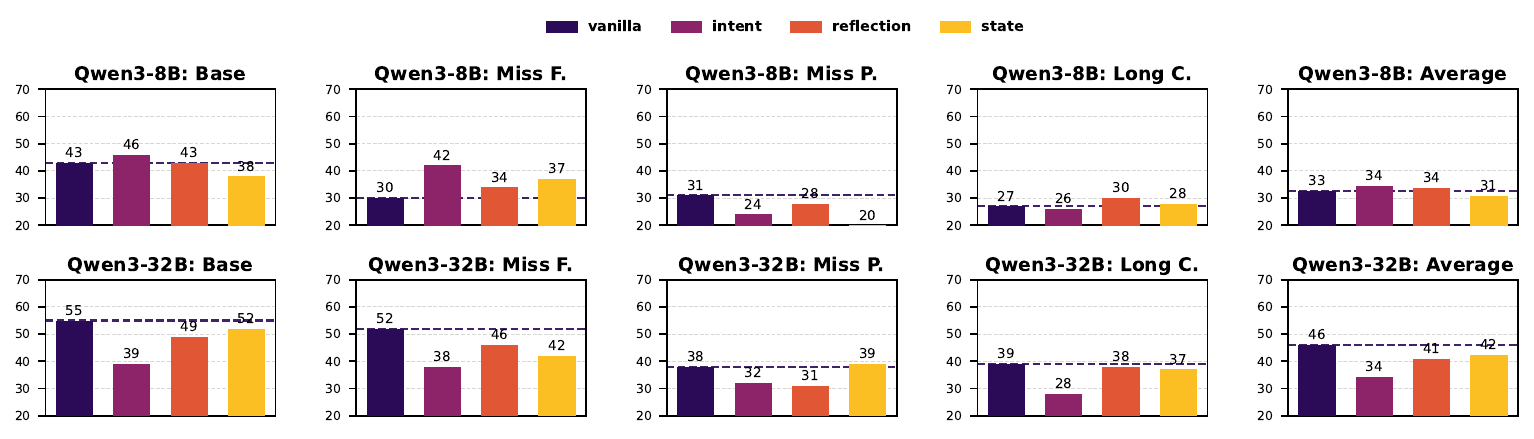}
    \caption{The Prompt-Hint results on BFCL-V3 dataset.}
    \label{fig:prompt_hint}
\end{figure*}

As shown in Figure~\ref{fig:prompt_hint}, prompt hints yield improvements in certain scenarios. However, the overall results indicate that such prompts often yield only limited gains and may even degrade tool-calling accuracy. A plausible explanation is that complex tool-use tasks involve multiple interacting factors, and \textbf{explicitly constraining the reasoning process to predefined perspectives can inadvertently restrict the model’s flexibility, causing it to overlook other critical information}.

\textbf{Width-based Parallel Sampling with Aggregation.}

Parallel sampling has proven effective in improving reasoning reliability across a wide range of LLM tasks \cite{DBLP:conf/iclr/0002WSLCNCZ23, DBLP:journals/corr/abs-2509-07980, DBLP:journals/corr/abs-2504-15466}. We extend this technique to multi-step tool execution and systematically evaluate its impact on tool-calling accuracy. Rather than generating an entire tool-call sequence in a single pass, we apply parallel sampling at each interaction step, where the model predicts the next action conditioned on the current dialogue history. At each step, multiple candidate actions are generated independently. If all candidates agree, the action is executed directly, otherwise, an aggregation function is applied to derive the final decision. The overall procedure is summarized in Algorithm~\ref{alg:parallel_multi_turn_tool_use}. We investigate two key factors: aggregation strategies and sampling scale.

\begin{figure}
    \centering
    \includegraphics[width=0.95\columnwidth]{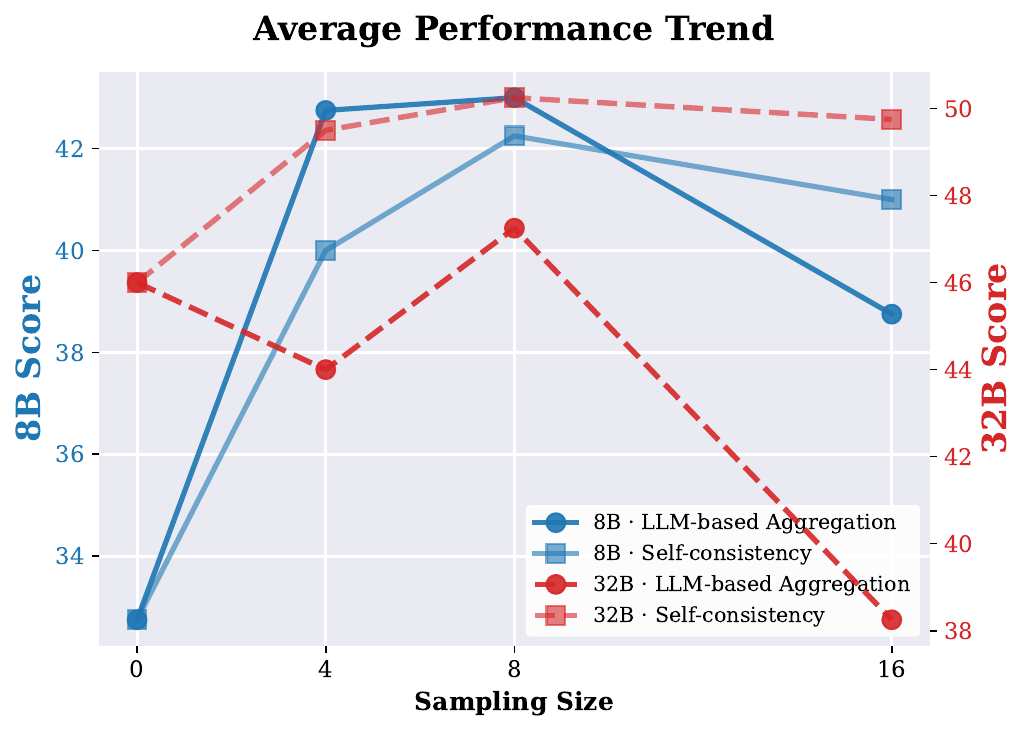}
    \caption{The average results of parallel sampling with different sampling size.}
    \label{fig:sampling_results}
\end{figure}
We apply and evaluate the following aggregation strategies $\mathcal{A(\cdot)}$:
(1) \textbf{Self-consistency} \cite{DBLP:conf/iclr/0002WSLCNCZ23}, which selects the final action by majority voting or consensus among parallel-sampled candidates;
(2) \textbf{LLM-based aggregation,} which feeds multiple sampled candidates back to the model to select the most appropriate action.

\begin{figure*}[tp]
    \centering
    \includegraphics[width=1\textwidth]{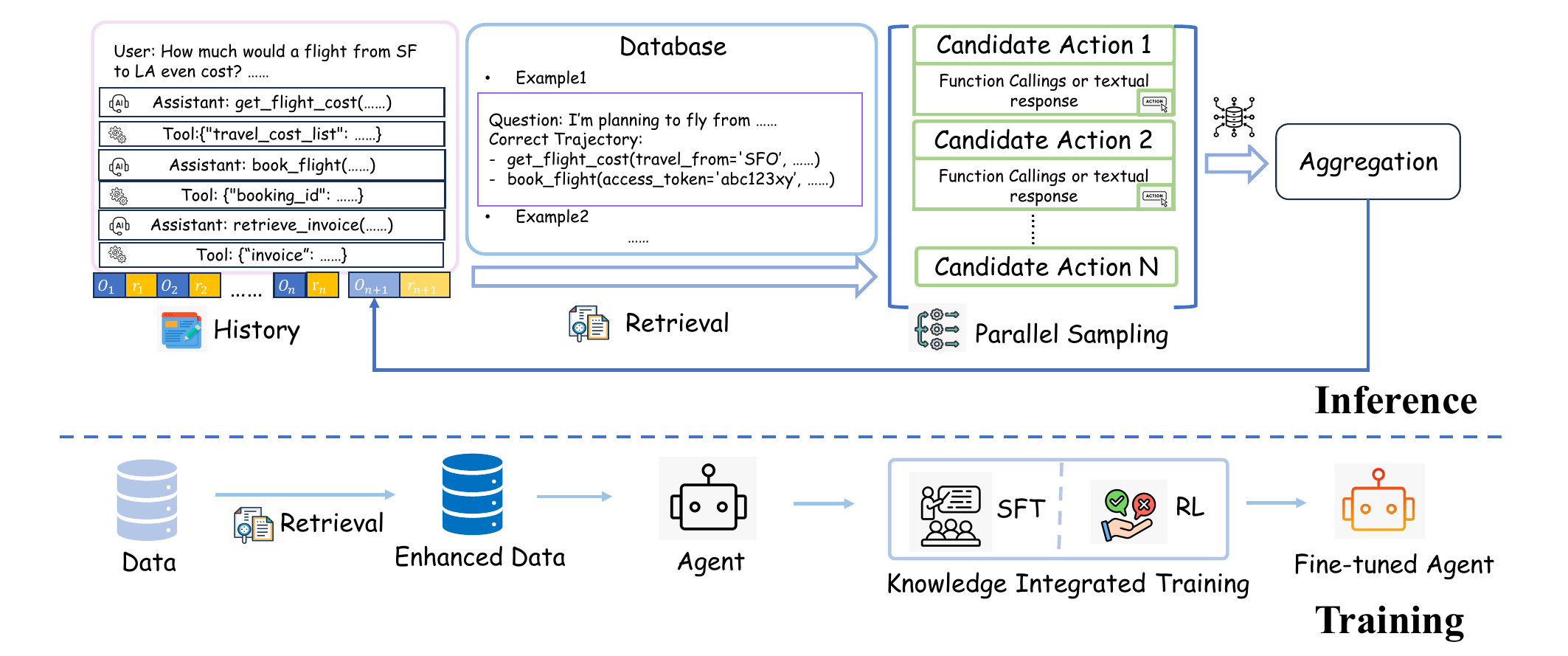}
    \caption{The inference and training framework of the method \textbf{KATE}.}
    \label{fig:frame}
\end{figure*}

As shown in Figure~\ref{fig:sampling_results}, results demonstrate that:
(1) \textbf{Effectiveness of parallel sampling.}
Parallel sampling substantially improves tool-calling performance, suggesting that greedy decoding with zero temperature fails to fully activate the model’s internal knowledge required for correct tool usage. Increasing the sampling temperature not only elicits relevant knowledge more effectively but also increases the frequency with which such knowledge appears, leading to more accurate tool invocation decisions.
(2) \textbf{Advantages of self-consistency.}
Self-consistency demonstrates greater stability and higher accuracy than LLM-based aggregation, with performance remaining largely insensitive to the sampling scale. This indicates that strong results can be achieved with a relatively small number of parallel samples. However, due to the diversity of textual outputs, self-consistency is less suitable for tasks requiring structured or diverse responses, such as code-based agent tool use \cite{DBLP:conf/acl/TrivediKHMDLGSB24}.
(3) \textbf{Performance of LLM-based aggregation.}
While LLM-based aggregation can further improve performance, its effectiveness does not increase monotonically with the number of parallel samples. This instability may stem from the model’s limited capacity to reason over long contexts containing many candidate actions.
These results suggest that expanding the width of the reasoning is more effective than encouraging more depth thinking. The completed results is illstruated in Figure~\ref{fig:sampling_results_all}.

\subsubsection{Training-Time}
We further investigate methodologies for internalizing knowledge into the model's parameters. We argue that inference-time reasoning enhancement alone is insufficient to fully improve a model’s tool-use capability, as knowledge injected through context is inherently limited. To achieve more robust gains, we further enhance tool-calling accuracy through post-training. Motivated by the hint-assisted reinforcement learning \cite{DBLP:journals/corr/abs-2505-16984, DBLP:journals/corr/abs-2504-14945}, we incorporate experiential knowledge directly into the training context by pre-inserting it as guidance signals. This design increases the probability of sampling correct reasoning trajectories in RL and thereby improving the overall efficacy of the training process. 

Our experimental framework spans the primary stages of the post-training regime, specifically SFT and RL. During the data preparation phase, we augment the training set by concatenating experiential knowledge retrieved from a structured knowledge base with the original user instructions to get the enhanced data. These augmented samples are then seamlessly integrated into the training pipeline. Through this design, we aim to quantify if there is the additive gain by incorporating training stage with provided experiential knowledge, analyzing its role in guiding the model toward more accurate tool selection.

\subsection{KATE}
Based on the above analysis, we propose our method \textbf{KATE}, which explicitly leverages knowledge across different stages of tool use. For the knowledge acquisition stage, we adopt Scenario-Trajectory (ST) knowledge to provide structured and reliable experience signals, as even the simple trajectory knowledge is effective. For the knowledge activation stage, we employ a depth-based parallel sampling and LLM-based aggregation strategy to effectively stimulate and utilize the acquired knowledge during reasoning. To further validate the effectiveness of training under our proposed knowledge-usage framework, we conduct additional experiments for training. These components demonstrate how knowledge systematically supports tool use at acquisition, activation, and training stages. The framework of our method is shown in Figure~\ref{fig:frame}. The inference process is in Algorithm~\ref{alg:stepwise_tool_use}.

\section{Experiment}
\begin{table*}[!ht]
    \centering
    \small
    \renewcommand{\arraystretch}{0.9}
    \setlength{\tabcolsep}{6pt}

    \begin{tabular}{l l r r r r r}
        \toprule
        \textbf{Model} & \textbf{Method} & \textbf{Base} & \textbf{Miss F.} & \textbf{Miss P.} & \textbf{Long C.} & \textbf{Average} \\
        \midrule

        \textbf{gpt-5} & FC & 49.0 & 35.0 & 30.0 & 37.0 & 37.75 \\
        \textbf{gpt-4.1} & FC & 52.0 & 39.0 & 36.0 & 50.0 & 44.25 \\
        \midrule

        \multirow{10}{*}{\textbf{Qwen3-8B}}
        & FC & 43.0 & 30.0 & 31.0 & 27.0 & 32.75 \\
        & Prompt & 38.0 & 38.0 & 28.0 & 17.0 & 30.25 \\
        & Memp & 52.0 & 25.0 & 36.0 & 31.0 & 36.00 \\
        \cmidrule{2-7}

        & \gcell{\textbf{KATE} (Ours)} &
        \gcell{59.0 \textcolor{blue}{\scriptsize (+16.0)}} &
        \gcell{41.0 \textcolor{blue}{\scriptsize (+11.0)}} &
        \gcell{41.0 \textcolor{blue}{\scriptsize (+10.0)}} &
        \gcell{40.0 \textcolor{blue}{\scriptsize (+13.0)}} &
        \gcell{46.00 \textcolor{blue}{\scriptsize (+13.25)}} \\

        & \gcell{$\quad \hookrightarrow$ w/o PS} &
        \gcell{49.0} & \gcell{35.0} & \gcell{38.0} & \gcell{30.0} & \gcell{38.00} \\

        & \gcell{$\quad \hookrightarrow$ w/o Exp} &
        \gcell{56.0} & \gcell{48.0} & \gcell{32.0} & \gcell{35.0} & \gcell{42.75} \\

        & \gcell{$\quad \hookrightarrow$ r PS-Con.} &
        \gcell{56.0} & \gcell{35.0} & \gcell{37.0} & \gcell{37.0} & \gcell{41.25} \\

        & $\quad \blacktriangleright$ + SFT &
        62.0 \textcolor{blue}{\scriptsize (+19.0)} &
        41.0 \textcolor{blue}{\scriptsize (+11.0)} &
        39.0 \textcolor{blue}{\scriptsize (+8.0)} &
        36.0 \textcolor{blue}{\scriptsize (+9.0)} &
        45.75 \textcolor{blue}{\scriptsize (+13.00)} \\

        & $\quad \blacktriangleright$ + RL &
        \textbf{64.0} \textcolor{blue}{\scriptsize (+21.0)} &
        43.0 \textcolor{blue}{\scriptsize (+13.0)} &
        \textbf{42.0} \textcolor{blue}{\scriptsize (+11.0)} &
        \textbf{42.0} \textcolor{blue}{\scriptsize (+15.0)} &
        \textbf{48.25} \textcolor{blue}{\scriptsize (+15.50)} \\

        & $\quad \blacktriangleright$ + SFT + RL &
        59.0 \textcolor{blue}{\scriptsize (+16.0)} &
        \textbf{45.0} \textcolor{blue}{\scriptsize (+15.0)} &
        39.0 \textcolor{blue}{\scriptsize (+8.0)} &
        \textbf{42.0} \textcolor{blue}{\scriptsize (+15.0)} &
        46.25 \textcolor{blue}{\scriptsize (+13.50)} \\

        \midrule

        \multirow{6}{*}{\textbf{Qwen3-32B}}
        & FC & 55.0 & 52.0 & 38.0 & 39.0 & 46.00 \\
        & Prompt & 55.0 & 47.0 & 42.0 & 37.0 & 45.25 \\
        & Memp & 63.0 & 44.0 & 47.0 & 42.0 & 49.00 \\
        \cmidrule{2-7}

        & \gcell{\textbf{KATE} (Ours)} &
        \gcell{62.0 \textcolor{blue}{\scriptsize (+7.0)}} &
        \gcell{\textbf{53.0} \textcolor{blue}{\scriptsize (+1.0)}} &
        \gcell{42.0 \textcolor{blue}{\scriptsize (+4.0)}} &
        \gcell{45.0 \textcolor{blue}{\scriptsize (+6.0)}} &
        \gcell{\textbf{50.50} \textcolor{blue}{\scriptsize (+4.50)}} \\

        & \gcell{$\quad \hookrightarrow$ w/o PS} &
        \gcell{62.0} & \gcell{43.0} & \gcell{48.0} & \gcell{43.0} & \gcell{49.00} \\

        & \gcell{$\quad \hookrightarrow$ w/o Exp} &
        \gcell{52.0} & \gcell{48.0} & \gcell{35.0} & \gcell{41.0} & \gcell{44.0} \\

        & \gcell{$\quad \hookrightarrow$ r PS-Con.} &
        \gcell{\textbf{65.0}} & \gcell{44.0} & \gcell{\textbf{46.0}} & \gcell{\textbf{46.0}} & \gcell{50.25} \\

        \bottomrule
    \end{tabular}

    \caption{Experimental results on BFCL-V3. Blue labels show absolute improvement over FC baseline. Green rows ($\hookrightarrow$) denote inference-time variants of ablation results, while non-highlighted rows ($\blacktriangleright$) indicate post-training results on Qwen3-8B. ``r PS-Con.'' means replace the LLM-based Aggregation with self-consistency.}
    \label{tab:bfcl_results}
\end{table*}

\subsection{Datasets}
We use BFCL-V3 \cite{DBLP:conf/icml/PatilMYJSSG25} and AppWorld \cite{DBLP:conf/acl/TrivediKHMDLGSB24} as our evaluation datasets. BFCL-V3 is a multi-step tool-use benchmark, which evaluates tool-use capability across diverse multi-turn interactive environments. Our study focuses on complex multi-turn interaction tasks spanning four scenarios in BFCL-V3, including \textit{Base}, \textit{Miss Func}, \textit{Miss Param}, and \textit{Long Context}. AppWorld is a benchmark of multi-step tasks for interactive coding agents, which use state-based programmatic evaluation approach. A task is successful if the final environment state matches the goal and all unit tests pass. It also provides two metrics: Task Goal Completion (TGC) and Scenario Goal Completion (SGC).

\subsection{Implementation Details}

In BFCL-V3 evaluation, we select 100 samples from the Base scenario as the training set, with the remaining data used for testing. To prevent data leakage, we partition the dataset by sample ID, using even-numbered instances for training and odd-numbered instances for testing. For AppWorld, we adopt a code-based tool-calling setting. Specifically, we distill the ground-truth solution procedures using GPT-4o, to obtain correct code-level reasoning steps to construct a knowledge base. Evaluation is conducted on test-normal (Test-N) and test-challenge (Test-C). The details are in Appendix~\ref{sec:appendix_inference_details}, and training details is in Appendix~\ref{sec:appendix_training_details}.

\begin{table}[t]
    \centering
    \small
    \renewcommand{\arraystretch}{0.9} 
    \setlength{\tabcolsep}{3pt}       
    \begin{tabular}{l r r r r r}
        \toprule
        & \multicolumn{2}{c}{\textbf{Test-N}} & \multicolumn{2}{c}{\textbf{Test-C}} & {\multirow{2}{*}{\textbf{Average}}} \\
        \cmidrule(lr){2-3} \cmidrule(lr){4-5}
        \multirow{-2}{*}{\textbf{Method}} & {\text{TGC}} & {\text{SGC}} & {\text{TGC}} & {\text{SGC}} & \\
        \midrule
        
        \multicolumn{6}{c}{\textbf{Qwen3-8B}} \\
        \cmidrule(lr){1-6} 
        ReAct & 10.1 & 1.8 & 3.8 & \textbf{0.7} & 4.1 \\
        ReAct + ST & 26.2 & \textbf{10.7} & 4.8 & \textbf{0.7} & 10.6 \\
        Memp & 22.0 & 7.1 & 3.6 & 0 & 7.92 \\
        \rowcolor[HTML]{F1F8E9} 
        \textbf{KATE} (Ours) & \textbf{26.8} & \textbf{10.7} & \textbf{5.5} & \textbf{0.7} & \textbf{10.92} \\
        
        \midrule
        
        \multicolumn{6}{c}{\textbf{Qwen3-32B}} \\
        \cmidrule(lr){1-6} 
        ReAct & 16.7 & 1.8 & 6.2 & \textbf{1.4} & 6.52 \\
        ReAct + ST & 27.4 & 1.8 & 8.6 & 0 & 9.45 \\
        Memp & 22.6 & 5.4 & \textbf{9.1} & \textbf{1.4} & 9.62 \\
        \rowcolor[HTML]{F1F8E9} 
        \textbf{KATE} (Ours) & \textbf{32.7} & \textbf{10.7} & 7.4 & 0.7 & \textbf{12.87} \\
        \bottomrule
    \end{tabular}
    \caption{Performance comparison on AppWorld.} 
    \label{tab:appworld_results}
\end{table}
\subsection{Baselines}
For dataset BFCL-V3, we adopt the following baselines: (1) Function Calling (FC) adopts the default tool-calling format. (2) Prompt-based methods (Prompt) use the default setting prompt in BFCL-V3 dataset. (3) Memp \cite{DBLP:journals/corr/abs-2508-06433} is a universal framework that enables AI agents to transform past task trajectories into reusable skills through the systematic management of procedural memory, while it achieves lifelong learning by continuously updating its trajectory repository, our experiments utilize a static, non-updating version for the purpose of fair comparison. For AppWorld, we adopt ReAct \cite{DBLP:conf/iclr/YaoZYDSN023} and Memp as the baseline.

\subsection{Result}
Table~\ref{tab:bfcl_results} presents the performance of various methods on the BFCL-V3 dataset. Both Qwen3-8B and Qwen3-32B show substantial improvements under our method, achieving roughly a performance gain over the baselines. We observe that our approach not only enhances performance on the \textit{Base scenario}, but also yields gains on \textit{Miss Func}, \textit{Miss Param}, and \textit{Long Context} tasks. By empowering models with explicit experiential knowledge, KATE even allows the Qwen3 series to outperform state-of-the-art models like GPT-4.1 and GPT-5 in specific tool-use benchmarks. Furthermore, fine-tuning confirms that internalizing knowledge into parameters provides benefits that exceed prompt-based injection alone.

As illustrated in Table~\ref{tab:appworld_results}, on AppWorld, KATE maintains a clear advantage over the ReAct baseline. However, in the Test-Challenge (Test-C) scenario, while KATE outperforms the vanilla ReAct and ReAct + ST, its improvement is slightly lower than Memp for some case. This stems from the extreme complexity of Test-C tasks: when a task exceeds the model’s inherent reasoning capacity, parallel sampling may fail to generate valid trajectories, and the presence of multiple candidate plans can introduce noise that interferes with final decision-making. This suggests a trade-off between reasoning width and task complexity. The reason is that for the too difficult question, model do not have the ability to answer, so the performance may have little lower than Memp in certain metrics. Importantly, this is not necessarily due to noise introduced by parallel sampling. Rather, for these hard tasks, the reasoning content produced by the model is generally incorrect, so aggregation tends to yield mostly noisy outputs. We believe that the most effective way to improve accuracy in such cases is either to provide additional training or supply higher-quality procedural knowledge to support the model during the reasoning process. The efficiency and more results are in Appendix~\ref{sec:appendix_inference_efficiency} and Appendix~\ref{sec:appendix_results}, respectively.

\subsection{Ablation Result}
As shown in Table~\ref{tab:bfcl_results}, ablation studies reveal critical insights into our components. ``w/o PS'' means without parallel sampling, it indicates that simply incorporating knowledge provides a baseline improvement. However, the gains are sightly lower than the full KATE framework. This suggests that without an activation mechanism, the model fails to fully utilize the injected knowledge. ``w/o Exp'' means without experiential knowledge. It shows that utilizing parallel sampling alone improves performance, but the model's upper bound remains limited. This indicates that task-specific experiential knowledge is essential for supplementing the model's inherent reasoning. Replacing LLM-based aggregation with self-consistency leads to gains on the Qwen3-32B model in some testing results, demonstrating the robustness of this strategy. 
\begin{figure}
    \centering
    \includegraphics[width=0.95\columnwidth]{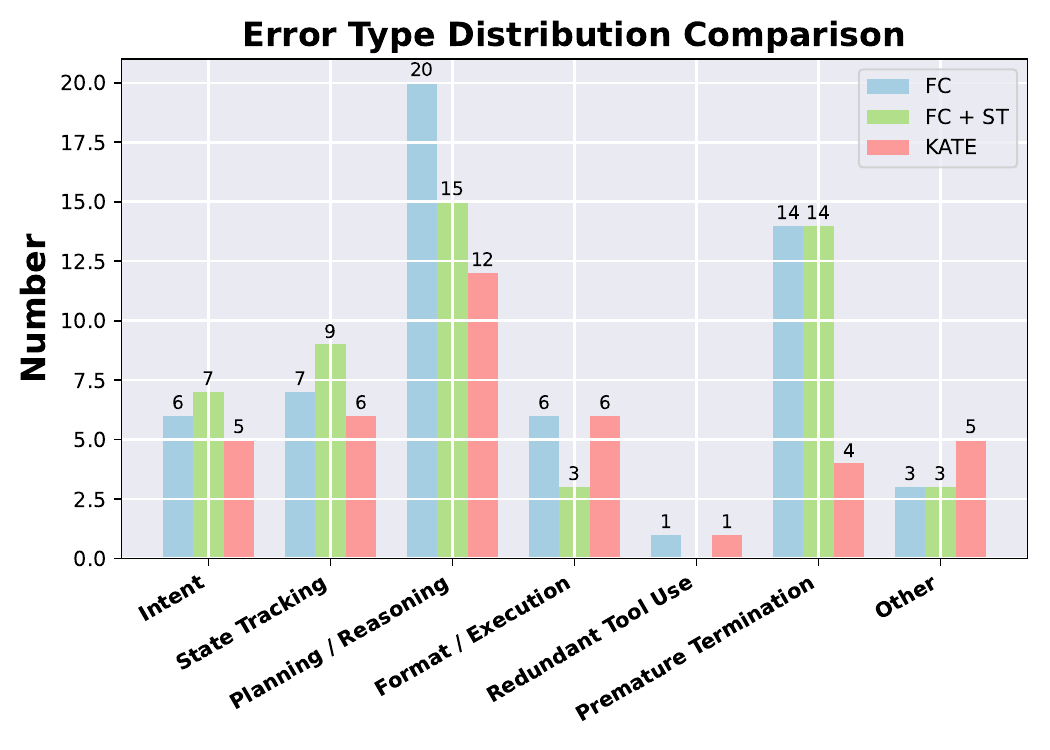}
    \caption{The Error type analysis of different methods on Qwen3-8B using dataset BFCL-V3 Base scenarios.}
    \label{fig:error_distribution}
\end{figure}

\subsection{Error Type Analysis}

As presented in Figure~\ref{fig:error_distribution}, we use gpt-5-mini to classify the error type. We find that planning and reasoning errors constitute the largest failure mode across all methods, but are substantially reduced by trajectory-level supervision and further by parallel sampling. This indicates that experiential knowledge can present the model to a broader set of previously observed scenarios, thereby improving its ability to reason and reducing reasoning-related failures. Parallel sampling also significantly mitigates premature termination errors, improving robustness in long-horizon execution.

\subsection{Training Analysis}

As shown in Table~\ref{tab:bfcl_results}, analysis of the Qwen3-8B fine-tuning experiments shows that RL is more effective than SFT for knowledge internalization. While the ``SFT + RL'' sequence is effective, we found that Direct RL (without prior SFT) yields the best performance. This suggests that for sufficiently strong base models, RL better explores and reinforces tool-calling capabilities than SFT within the same data budget. As shown in Figure \ref{fig:curves}, while the difference between ``SFT + RL'' and ``RL'' is subtle, RL consistently maintains an upper hand in convergence quality and final accuracy.

\section{Related Work}
\textbf{Tool Learning.} LLMs have recently been extended with tool-use capabilities \cite{DBLP:conf/aaai/HaoCJL0LZ25, DBLP:journals/corr/abs-2504-03601, DBLP:conf/iclr/QinLYZYLLCTQZHT24}, enabling them to interact with external APIs and environments beyond pure text generation \cite{DBLP:journals/corr/abs-2510-10197}. Due to the complexity of multi-turn interactive tool-use tasks, researchers typically enhance tool-learning capabilities by optimizing reasoning frameworks \cite{DBLP:conf/iclr/QinLYZYLLCTQZHT24} and fine-tuning model parameters \cite{DBLP:journals/corr/abs-2510-10197}. With the adoption of reinforcement learning in LLMs \cite{DBLP:journals/corr/abs-2501-03262, DBLP:journals/corr/abs-2402-03300}, an increasing number of tasks leverage RL to strengthen a model’s ability to invoke tools \cite{DBLP:journals/corr/abs-2504-13958, DBLP:journals/corr/abs-2509-01055, DBLP:journals/corr/abs-2509-02479}. However, few studies have emphasized the critical role of knowledge in tool-use tasks.  

\textbf{Experiential Knowledge.} Experiential knowledge refers to the experiences, memories, and thought processes involved in deriving an answer \cite{DBLP:journals/corr/abs-2508-06433, DBLP:journals/corr/abs-2508-19005}. By applying such knowledge to downstream tasks, models are provided with experiential guidance for similar scenarios, thereby facilitating correct responses \cite{cao2025remembermerefineme, DBLP:journals/corr/abs-2507-06229, DBLP:conf/iclr/ZhengWW024, DBLP:journals/corr/abs-2507-23361, DBLP:journals/corr/abs-2504-19413, DBLP:conf/icml/WangMFN25}. Increasingly, recent methods leverage procedural experience to enhance model capabilities during both reasoning and training stages. But they don't study the role of experiential knowledge in different stage of tool use, we systematically investigate the full lifecycle of knowledge in tool learning. At the acquisition stage, prior works~\cite{cao2025remembermerefineme, DBLP:journals/corr/abs-2508-06433, DBLP:conf/icml/WangMFN25, DBLP:journals/corr/abs-2508-16153, DBLP:journals/corr/abs-2507-06229, DBLP:journals/corr/abs-2508-19005} often focus on improving specific knowledge construction process. In contrast, we introduce and compare knowledge at different levels of abstraction (instance-level trajectories and intent-level scripts), and analyze their distinct effects. This allows us to understand which type of knowledge is most actionable for tool-using agents. At activation stage, instead of designing more fine-grained retrieval mechanisms like previous works~\cite{cao2025remembermerefineme, DBLP:journals/corr/abs-2508-16153} to retrieve most useful knowledge, we ask how already acquired knowledge should be utilized during inference with a simple top-k retriever. 

\section{Conclusion}
In this work, we investigate how do knowledge influence LLMs in multi-turn tool-use tasks. We categorize experiential knowledge into instance-level and intent-level forms and systematically evaluate their impact on tool execution. We further study how such knowledge can be effectively activated during inference and find that increasing reasoning breadth is particularly effective in eliciting latent experiential knowledge. We additionally fine-tune the model to further consolidate knowledge-grounded reasoning. By integrating knowledge and tool use across the stages of knowledge construction, inference-time activation, and training-time refinement, we demonstrate how experiential knowledge can systematically enhance tool execution capabilities.

\section*{Limitations}
Our experiments demonstrate that incorporating knowledge can effectively enhance tool-calling performance, and that explicit knowledge activation further improves tool-use accuracy. Nevertheless, our evaluation is conducted on a relatively small-scale knowledge base, and the impact of scaling the knowledge repository remains unexplored. Moreover, our current study is limited to text-only tool-use scenarios, leaving the extension to multimodal tasks as an important direction for future work.

\section*{Ethics Statement}
Our work does not introduce ethical concerns. This paper utilized AI assistance for language polishing of the manuscript, including vocabulary correction and spell checking. 

\section*{Acknowledgments}
This work is supported by the National Natural Science Foundation of China (No. U24A20335).

\bibliography{custom}

\begin{thebibliography}{38}
\providecommand{\natexlab}[1]{#1}

\bibitem[{Cai et~al.(2025)Cai, Hao, Zhou, Yan, Lei, Zhen, Han, Yang, Li, Pan, Huai, Chen, Li, Chen, Zhang, Qiu, and He}]{DBLP:journals/corr/abs-2508-19005}
Yuxuan Cai, Yipeng Hao, Jie Zhou, Hang Yan, Zhikai Lei, Rui Zhen, Zhenhua Han, Yutao Yang, Junsong Li, Qianjun Pan, Tianyu Huai, Qin Chen, Xin Li, Kai Chen, Bo~Zhang, Xipeng Qiu, and Liang He. 2025.
\newblock \href {https://doi.org/10.48550/ARXIV.2508.19005} {Building self-evolving agents via experience-driven lifelong learning: {A} framework and benchmark}.
\newblock \emph{CoRR}, abs/2508.19005.

\bibitem[{Cao et~al.(2025)Cao, Deng, Yu, Zhou, Liu, Ding, and Zhao}]{cao2025remembermerefineme}
Zouying Cao, Jiaji Deng, Li~Yu, Weikang Zhou, Zhaoyang Liu, Bolin Ding, and Hai Zhao. 2025.
\newblock \href {https://arxiv.org/abs/2512.10696} {Remember me, refine me: A dynamic procedural memory framework for experience-driven agent evolution}.
\newblock \emph{Preprint}, arXiv:2512.10696.

\bibitem[{Chen et~al.(2025)Chen, Lin, Gu, Shi, Lian, Yun, Chen, Sun, Cao, and Wang}]{DBLP:journals/corr/abs-2507-23361}
Silin Chen, Shaoxin Lin, Xiaodong Gu, Yuling Shi, Heng Lian, Longfei Yun, Dong Chen, Weiguo Sun, Lin Cao, and Qianxiang Wang. 2025.
\newblock \href {https://doi.org/10.48550/ARXIV.2507.23361} {Swe-exp: Experience-driven software issue resolution}.
\newblock \emph{CoRR}, abs/2507.23361.

\bibitem[{Chhikara et~al.(2025)Chhikara, Khant, Aryan, Singh, and Yadav}]{DBLP:journals/corr/abs-2504-19413}
Prateek Chhikara, Dev Khant, Saket Aryan, Taranjeet Singh, and Deshraj Yadav. 2025.
\newblock \href {https://doi.org/10.48550/ARXIV.2504.19413} {Mem0: Building production-ready {AI} agents with scalable long-term memory}.
\newblock \emph{CoRR}, abs/2504.19413.

\bibitem[{Fang et~al.(2025)Fang, Liang, Wang, Wu, Qiao, Xie, Huang, Chen, and Zhang}]{DBLP:journals/corr/abs-2508-06433}
Runnan Fang, Yuan Liang, Xiaobin Wang, Jialong Wu, Shuofei Qiao, Pengjun Xie, Fei Huang, Huajun Chen, and Ningyu Zhang. 2025.
\newblock \href {https://doi.org/10.48550/ARXIV.2508.06433} {Memp: Exploring agent procedural memory}.
\newblock \emph{CoRR}, abs/2508.06433.

\bibitem[{Hao et~al.(2025)Hao, Cao, Jin, Liao, Chen, Liu, and Zhao}]{DBLP:conf/aaai/HaoCJL0LZ25}
Yupu Hao, Pengfei Cao, Zhuoran Jin, Huanxuan Liao, Yubo Chen, Kang Liu, and Jun Zhao. 2025.
\newblock \href {https://doi.org/10.1609/AAAI.V39I22.34573} {{CITI:} enhancing tool utilizing ability in large language models without sacrificing general performance}.
\newblock In \emph{AAAI-25, Sponsored by the Association for the Advancement of Artificial Intelligence, February 25 - March 4, 2025, Philadelphia, PA, {USA}}, pages 23996--24004. {AAAI} Press.

\bibitem[{Hu(2025)}]{DBLP:journals/corr/abs-2501-03262}
Jian Hu. 2025.
\newblock \href {https://doi.org/10.48550/ARXIV.2501.03262} {{REINFORCE++:} {A} simple and efficient approach for aligning large language models}.
\newblock \emph{CoRR}, abs/2501.03262.

\bibitem[{Jiang et~al.(2025)Jiang, Lu, Li, Lyu, Nie, Wang, Su, Chen, Zou, Du, Pang, and Chen}]{DBLP:journals/corr/abs-2509-01055}
Dongfu Jiang, Yi~Lu, Zhuofeng Li, Zhiheng Lyu, Ping Nie, Haozhe Wang, Alex Su, Hui Chen, Kai Zou, Chao Du, Tianyu Pang, and Wenhu Chen. 2025.
\newblock \href {https://doi.org/10.48550/ARXIV.2509.01055} {Verltool: Towards holistic agentic reinforcement learning with tool use}.
\newblock \emph{CoRR}, abs/2509.01055.

\bibitem[{Jin et~al.(2025)Jin, Zeng, Yue, Wang, Zamani, and Han}]{DBLP:journals/corr/abs-2503-09516}
Bowen Jin, Hansi Zeng, Zhenrui Yue, Dong Wang, Hamed Zamani, and Jiawei Han. 2025.
\newblock \href {https://doi.org/10.48550/ARXIV.2503.09516} {Search-r1: Training llms to reason and leverage search engines with reinforcement learning}.
\newblock \emph{CoRR}, abs/2503.09516.

\bibitem[{Li et~al.()Li, Jin, Men, Hao, Zhu, Wang, Huang, Wang, Hua, Wang et~al.}]{liagentic}
Jiachun Li, Zhuoran Jin, Tianyi Men, Yupu Hao, Kejian Zhu, Lingshuai Wang, Dongqi Huang, Longxiang Wang, Shengjia Hua, Lu~Wang, and 1 others.
\newblock Agentic environment engineering for large language models: A survey of environment modeling, synthesis, evaluation, and application.

\bibitem[{Li et~al.(2023)Li, Zhao, Yu, Song, Li, Yu, Li, Huang, and Li}]{DBLP:conf/emnlp/LiZ000YLHL23}
Minghao Li, Yingxiu Zhao, Bowen Yu, Feifan Song, Hangyu Li, Haiyang Yu, Zhoujun Li, Fei Huang, and Yongbin Li. 2023.
\newblock \href {https://doi.org/10.18653/V1/2023.EMNLP-MAIN.187} {Api-bank: {A} comprehensive benchmark for tool-augmented llms}.
\newblock In \emph{Proceedings of the 2023 Conference on Empirical Methods in Natural Language Processing, {EMNLP} 2023, Singapore, December 6-10, 2023}, pages 3102--3116. Association for Computational Linguistics.

\bibitem[{Li et~al.(2025)Li, Zou, and Liu}]{DBLP:journals/corr/abs-2503-23383}
Xuefeng Li, Haoyang Zou, and Pengfei Liu. 2025.
\newblock \href {https://doi.org/10.48550/ARXIV.2503.23383} {Torl: Scaling tool-integrated {RL}}.
\newblock \emph{CoRR}, abs/2503.23383.

\bibitem[{Liu et~al.(2025{\natexlab{a}})Liu, Farina, and Ozdaglar}]{DBLP:journals/corr/abs-2505-16984}
Mingyang Liu, Gabriele Farina, and Asuman~E. Ozdaglar. 2025{\natexlab{a}}.
\newblock \href {https://doi.org/10.48550/ARXIV.2505.16984} {{UFT:} unifying supervised and reinforcement fine-tuning}.
\newblock \emph{CoRR}, abs/2505.16984.

\bibitem[{Liu et~al.(2025{\natexlab{b}})Liu, Huang, Zeng, Hao, Yu, Li, Wang, Gan, Liu, Yu, Wang, Wang, Ning, Hou, Wang, Wu, Wang, Liu, Wang, Tang, Tu, Shang, Jiang, Tang, Lian, Liu, and Chen}]{DBLP:conf/iclr/Liu0ZHYL0GLY0WN25}
Weiwen Liu, Xu~Huang, Xingshan Zeng, Xinlong Hao, Shuai Yu, Dexun Li, Shuai Wang, Weinan Gan, Zhengying Liu, Yuanqing Yu, Zezhong Wang, Yuxian Wang, Wu~Ning, Yutai Hou, Bin Wang, Chuhan Wu, Xinzhi Wang, Yong Liu, Yasheng Wang, and 8 others. 2025{\natexlab{b}}.
\newblock \href {https://openreview.net/forum?id=8EB8k6DdCU} {Toolace: Winning the points of {LLM} function calling}.
\newblock In \emph{The Thirteenth International Conference on Learning Representations, {ICLR} 2025, Singapore, April 24-28, 2025}. OpenReview.net.

\bibitem[{Lu et~al.(2025)Lu, Wang, Zhang, Wu, Gan, Zhuang, Gu, and Lin}]{DBLP:journals/corr/abs-2510-10197}
Siyuan Lu, Zechuan Wang, Hongxuan Zhang, Qintong Wu, Leilei Gan, Chenyi Zhuang, Jinjie Gu, and Tao Lin. 2025.
\newblock \href {https://doi.org/10.48550/ARXIV.2510.10197} {Don't just fine-tune the agent, tune the environment}.
\newblock \emph{CoRR}, abs/2510.10197.

\bibitem[{Mialon et~al.(2023)Mialon, Dess{\`{\i}}, Lomeli, Nalmpantis, Pasunuru, Raileanu, Rozi{\`{e}}re, Schick, Dwivedi{-}Yu, Celikyilmaz, Grave, LeCun, and Scialom}]{DBLP:journals/tmlr/MialonDLNPRRSDC23}
Gr{\'{e}}goire Mialon, Roberto Dess{\`{\i}}, Maria Lomeli, Christoforos Nalmpantis, Ramakanth Pasunuru, Roberta Raileanu, Baptiste Rozi{\`{e}}re, Timo Schick, Jane Dwivedi{-}Yu, Asli Celikyilmaz, Edouard Grave, Yann LeCun, and Thomas Scialom. 2023.
\newblock \href {https://openreview.net/forum?id=jh7wH2AzKK} {Augmented language models: a survey}.
\newblock \emph{Trans. Mach. Learn. Res.}, 2023.

\bibitem[{Pan et~al.(2025)Pan, Li, Lian, Snell, Zhou, Yala, Darrell, Keutzer, and Suhr}]{DBLP:journals/corr/abs-2504-15466}
Jiayi Pan, Xiuyu Li, Long Lian, Charlie Snell, Yifei Zhou, Adam Yala, Trevor Darrell, Kurt Keutzer, and Alane Suhr. 2025.
\newblock \href {https://doi.org/10.48550/ARXIV.2504.15466} {Learning adaptive parallel reasoning with language models}.
\newblock \emph{CoRR}, abs/2504.15466.

\bibitem[{Patil et~al.(2025)Patil, Mao, Yan, Ji, Suresh, Stoica, and Gonzalez}]{DBLP:conf/icml/PatilMYJSSG25}
Shishir~G. Patil, Huanzhi Mao, Fanjia Yan, Charlie~Cheng{-}Jie Ji, Vishnu Suresh, Ion Stoica, and Joseph~E. Gonzalez. 2025.
\newblock \href {https://openreview.net/forum?id=2GmDdhBdDk} {The berkeley function calling leaderboard {(BFCL):} from tool use to agentic evaluation of large language models}.
\newblock In \emph{Forty-second International Conference on Machine Learning, {ICML} 2025, Vancouver, BC, Canada, July 13-19, 2025}. OpenReview.net.

\bibitem[{Plaat et~al.(2025)Plaat, van Duijn, van Stein, Preuss, van~der Putten, and Batenburg}]{DBLP:journals/corr/abs-2503-23037}
Aske Plaat, Max~J. van Duijn, Niki van Stein, Mike Preuss, Peter van~der Putten, and Kees~Joost Batenburg. 2025.
\newblock \href {https://doi.org/10.48550/ARXIV.2503.23037} {Agentic large language models, a survey}.
\newblock \emph{CoRR}, abs/2503.23037.

\bibitem[{Prabhakar et~al.(2025)Prabhakar, Liu, Zhu, Zhang, Awalgaonkar, Wang, Liu, Chen, Hoang, Niebles, Heinecke, Yao, Wang, Savarese, and Xiong}]{DBLP:journals/corr/abs-2504-03601}
Akshara Prabhakar, Zuxin Liu, Ming Zhu, Jianguo Zhang, Tulika Awalgaonkar, Shiyu Wang, Zhiwei Liu, Haolin Chen, Thai Hoang, Juan~Carlos Niebles, Shelby Heinecke, Weiran Yao, Huan Wang, Silvio Savarese, and Caiming Xiong. 2025.
\newblock \href {https://doi.org/10.48550/ARXIV.2504.03601} {Apigen-mt: Agentic pipeline for multi-turn data generation via simulated agent-human interplay}.
\newblock \emph{CoRR}, abs/2504.03601.

\bibitem[{Qian et~al.(2025)Qian, Acikgoz, He, Wang, Chen, Hakkani{-}T{\"{u}}r, Tur, and Ji}]{DBLP:journals/corr/abs-2504-13958}
Cheng Qian, Emre~Can Acikgoz, Qi~He, Hongru Wang, Xiusi Chen, Dilek Hakkani{-}T{\"{u}}r, Gokhan Tur, and Heng Ji. 2025.
\newblock \href {https://doi.org/10.48550/ARXIV.2504.13958} {Toolrl: Reward is all tool learning needs}.
\newblock \emph{CoRR}, abs/2504.13958.

\bibitem[{Qin et~al.(2024)Qin, Liang, Ye, Zhu, Yan, Lu, Lin, Cong, Tang, Qian, Zhao, Hong, Tian, Xie, Zhou, Gerstein, Li, Liu, and Sun}]{DBLP:conf/iclr/QinLYZYLLCTQZHT24}
Yujia Qin, Shihao Liang, Yining Ye, Kunlun Zhu, Lan Yan, Yaxi Lu, Yankai Lin, Xin Cong, Xiangru Tang, Bill Qian, Sihan Zhao, Lauren Hong, Runchu Tian, Ruobing Xie, Jie Zhou, Mark Gerstein, Dahai Li, Zhiyuan Liu, and Maosong Sun. 2024.
\newblock \href {https://openreview.net/forum?id=dHng2O0Jjr} {Toolllm: Facilitating large language models to master 16000+ real-world apis}.
\newblock In \emph{The Twelfth International Conference on Learning Representations, {ICLR} 2024, Vienna, Austria, May 7-11, 2024}. OpenReview.net.

\bibitem[{Qu et~al.(2025)Qu, Dai, Wei, Cai, Wang, Yin, Xu, and Wen}]{DBLP:conf/iclr/QuDWCWY0W25}
Changle Qu, Sunhao Dai, Xiaochi Wei, Hengyi Cai, Shuaiqiang Wang, Dawei Yin, Jun Xu, and Ji{-}Rong Wen. 2025.
\newblock \href {https://openreview.net/forum?id=QKBu1BOAwd} {From exploration to mastery: Enabling llms to master tools via self-driven interactions}.
\newblock In \emph{The Thirteenth International Conference on Learning Representations, {ICLR} 2025, Singapore, April 24-28, 2025}. OpenReview.net.

\bibitem[{Shao et~al.(2024)Shao, Wang, Zhu, Xu, Song, Zhang, Li, Wu, and Guo}]{DBLP:journals/corr/abs-2402-03300}
Zhihong Shao, Peiyi Wang, Qihao Zhu, Runxin Xu, Junxiao Song, Mingchuan Zhang, Y.~K. Li, Y.~Wu, and Daya Guo. 2024.
\newblock \href {https://doi.org/10.48550/ARXIV.2402.03300} {Deepseekmath: Pushing the limits of mathematical reasoning in open language models}.
\newblock \emph{CoRR}, abs/2402.03300.

\bibitem[{Shinn et~al.(2023)Shinn, Cassano, Gopinath, Narasimhan, and Yao}]{DBLP:conf/nips/ShinnCGNY23}
Noah Shinn, Federico Cassano, Ashwin Gopinath, Karthik Narasimhan, and Shunyu Yao. 2023.
\newblock \href {http://papers.nips.cc/paper\_files/paper/2023/hash/1b44b878bb782e6954cd888628510e90-Abstract-Conference.html} {Reflexion: language agents with verbal reinforcement learning}.
\newblock In \emph{Advances in Neural Information Processing Systems 36: Annual Conference on Neural Information Processing Systems 2023, NeurIPS 2023, New Orleans, LA, USA, December 10 - 16, 2023}.

\bibitem[{Tang et~al.(2025)Tang, Qin, Peng, Zhou, Shao, Du, Wei, Xia, Wu, Zhu, Zhang, Liu, Wang, Hong, Wu, Cheng, Wang, and Zhou}]{DBLP:journals/corr/abs-2507-06229}
Xiangru Tang, Tianrui Qin, Tianhao Peng, Ziyang Zhou, Daniel Shao, Tingting Du, Xinming Wei, Peng Xia, Fang Wu, He~Zhu, Ge~Zhang, Jiaheng Liu, Xingyao Wang, Sirui Hong, Chenglin Wu, Hao Cheng, Chi Wang, and Wangchunshu Zhou. 2025.
\newblock \href {https://doi.org/10.48550/ARXIV.2507.06229} {Agent {KB:} leveraging cross-domain experience for agentic problem solving}.
\newblock \emph{CoRR}, abs/2507.06229.

\bibitem[{Trivedi et~al.(2024)Trivedi, Khot, Hartmann, Manku, Dong, Li, Gupta, Sabharwal, and Balasubramanian}]{DBLP:conf/acl/TrivediKHMDLGSB24}
Harsh Trivedi, Tushar Khot, Mareike Hartmann, Ruskin Manku, Vinty Dong, Edward Li, Shashank Gupta, Ashish Sabharwal, and Niranjan Balasubramanian. 2024.
\newblock \href {https://doi.org/10.18653/V1/2024.ACL-LONG.850} {Appworld: {A} controllable world of apps and people for benchmarking interactive coding agents}.
\newblock In \emph{Proceedings of the 62nd Annual Meeting of the Association for Computational Linguistics (Volume 1: Long Papers), {ACL} 2024, Bangkok, Thailand, August 11-16, 2024}, pages 16022--16076. Association for Computational Linguistics.

\bibitem[{Wang et~al.(2024)Wang, Yao, Xu, Qiao, Deng, Wang, Chen, Gu, Jiang, Xie, Huang, Chen, and Zhang}]{DBLP:conf/emnlp/WangYXQD00GJX0C24}
Mengru Wang, Yunzhi Yao, Ziwen Xu, Shuofei Qiao, Shumin Deng, Peng Wang, Xiang Chen, Jia{-}Chen Gu, Yong Jiang, Pengjun Xie, Fei Huang, Huajun Chen, and Ningyu Zhang. 2024.
\newblock \href {https://doi.org/10.18653/V1/2024.FINDINGS-EMNLP.416} {Knowledge mechanisms in large language models: {A} survey and perspective}.
\newblock In \emph{Findings of the Association for Computational Linguistics: {EMNLP} 2024, Miami, Florida, USA, November 12-16, 2024}, pages 7097--7135. Association for Computational Linguistics.

\bibitem[{Wang et~al.(2023)Wang, Wei, Schuurmans, Le, Chi, Narang, Chowdhery, and Zhou}]{DBLP:conf/iclr/0002WSLCNCZ23}
Xuezhi Wang, Jason Wei, Dale Schuurmans, Quoc~V. Le, Ed~H. Chi, Sharan Narang, Aakanksha Chowdhery, and Denny Zhou. 2023.
\newblock \href {https://openreview.net/forum?id=1PL1NIMMrw} {Self-consistency improves chain of thought reasoning in language models}.
\newblock In \emph{The Eleventh International Conference on Learning Representations, {ICLR} 2023, Kigali, Rwanda, May 1-5, 2023}. OpenReview.net.

\bibitem[{Wang et~al.(2025)Wang, Mao, Fried, and Neubig}]{DBLP:conf/icml/WangMFN25}
Zora~Zhiruo Wang, Jiayuan Mao, Daniel Fried, and Graham Neubig. 2025.
\newblock \href {https://openreview.net/forum?id=NTAhi2JEEE} {Agent workflow memory}.
\newblock In \emph{Forty-second International Conference on Machine Learning, {ICML} 2025, Vancouver, BC, Canada, July 13-19, 2025}. OpenReview.net.

\bibitem[{Xue et~al.(2025)Xue, Zheng, Liu, Li, Zheng, Ma, and An}]{DBLP:journals/corr/abs-2509-02479}
Zhenghai Xue, Longtao Zheng, Qian Liu, Yingru Li, Xiaosen Zheng, Zejun Ma, and Bo~An. 2025.
\newblock \href {https://doi.org/10.48550/ARXIV.2509.02479} {Simpletir: End-to-end reinforcement learning for multi-turn tool-integrated reasoning}.
\newblock \emph{CoRR}, abs/2509.02479.

\bibitem[{Yan et~al.(2025)Yan, Li, Hu, Wang, Cui, Qu, Cheng, and Zhang}]{DBLP:journals/corr/abs-2504-14945}
Jianhao Yan, Yafu Li, Zican Hu, Zhi Wang, Ganqu Cui, Xiaoye Qu, Yu~Cheng, and Yue Zhang. 2025.
\newblock \href {https://doi.org/10.48550/ARXIV.2504.14945} {Learning to reason under off-policy guidance}.
\newblock \emph{CoRR}, abs/2504.14945.

\bibitem[{Yang et~al.(2025)Yang, Li, Yang, Zhang, Hui, Zheng, Yu, Gao, Huang, Lv, Zheng, Liu, Zhou, Huang, Hu, Ge, Wei, Lin, Tang, Yang, Tu, Zhang, Yang, Yang, Zhou, Lin, Dang, Bao, Yang, Yu, Deng, Li, Xue, Li, Zhang, Wang, Zhu, Men, Gao, Liu, Luo, Li, Tang, Yin, Ren, Wang, Zhang, Ren, Fan, Su, Zhang, Zhang, Wan, Liu, Wang, Cui, Zhang, Zhou, and Qiu}]{DBLP:journals/corr/abs-2505-09388}
An~Yang, Anfeng Li, Baosong Yang, Beichen Zhang, Binyuan Hui, Bo~Zheng, Bowen Yu, Chang Gao, Chengen Huang, Chenxu Lv, Chujie Zheng, Dayiheng Liu, Fan Zhou, Fei Huang, Feng Hu, Hao Ge, Haoran Wei, Huan Lin, Jialong Tang, and 40 others. 2025.
\newblock \href {https://doi.org/10.48550/ARXIV.2505.09388} {Qwen3 technical report}.
\newblock \emph{CoRR}, abs/2505.09388.

\bibitem[{Yao et~al.(2023)Yao, Zhao, Yu, Du, Shafran, Narasimhan, and Cao}]{DBLP:conf/iclr/YaoZYDSN023}
Shunyu Yao, Jeffrey Zhao, Dian Yu, Nan Du, Izhak Shafran, Karthik~R. Narasimhan, and Yuan Cao. 2023.
\newblock \href {https://openreview.net/forum?id=WE\_vluYUL-X} {React: Synergizing reasoning and acting in language models}.
\newblock In \emph{The Eleventh International Conference on Learning Representations, {ICLR} 2023, Kigali, Rwanda, May 1-5, 2023}. OpenReview.net.

\bibitem[{Zhang et~al.(2025)Zhang, Dong, Zhang, Kautz, Catanzaro, Tao, Wu, Yu, and Liu}]{DBLP:journals/corr/abs-2505-00024}
Shaokun Zhang, Yi~Dong, Jieyu Zhang, Jan Kautz, Bryan Catanzaro, Andrew Tao, Qingyun Wu, Zhiding Yu, and Guilin Liu. 2025.
\newblock \href {https://doi.org/10.48550/ARXIV.2505.00024} {Nemotron-research-tool-n1: Exploring tool-using language models with reinforced reasoning}.
\newblock \emph{CoRR}, abs/2505.00024.

\bibitem[{Zheng et~al.(2024)Zheng, Wang, Wang, and An}]{DBLP:conf/iclr/ZhengWW024}
Longtao Zheng, Rundong Wang, Xinrun Wang, and Bo~An. 2024.
\newblock \href {https://openreview.net/forum?id=Pc8AU1aF5e} {Synapse: Trajectory-as-exemplar prompting with memory for computer control}.
\newblock In \emph{The Twelfth International Conference on Learning Representations, {ICLR} 2024, Vienna, Austria, May 7-11, 2024}. OpenReview.net.

\bibitem[{Zheng et~al.(2025)Zheng, Zhang, Yu, Wang, Dai, Liu, Bao, Huang, Huang, and Yu}]{DBLP:journals/corr/abs-2509-07980}
Tong Zheng, Hongming Zhang, Wenhao Yu, Xiaoyang Wang, Runpeng Dai, Rui Liu, Huiwen Bao, Chengsong Huang, Heng Huang, and Dong Yu. 2025.
\newblock \href {https://doi.org/10.48550/ARXIV.2509.07980} {Parallel-r1: Towards parallel thinking via reinforcement learning}.
\newblock \emph{CoRR}, abs/2509.07980.

\bibitem[{Zhou et~al.(2025)Zhou, Chen, Guo, Yan, Lee, Wang, Lee, Zhang, Shao, Yang, and Wang}]{DBLP:journals/corr/abs-2508-16153}
Huichi Zhou, Yihang Chen, Siyuan Guo, Xue Yan, Kin~Hei Lee, Zihan Wang, Ka~Yiu Lee, Guchun Zhang, Kun Shao, Linyi Yang, and Jun Wang. 2025.
\newblock \href {https://doi.org/10.48550/ARXIV.2508.16153} {Memento: Fine-tuning {LLM} agents without fine-tuning llms}.
\newblock \emph{CoRR}, abs/2508.16153.

\end{thebibliography}

\appendix
\newpage
\section{Knowledge Construction for Augmentation}
\label{sec:appendix_knowledge_construction}
We use GPT-4o to summarize the experiential knowledge. 

We construct experiential data in the following manner. For \textbf{Scenario Trajectory Knowledge (ST)}, we directly use the ground-truth tool invocation list from the training data as experiential knowledge. For \textbf{Experience Summary Knowledge (ES)}, we ask the LLM to generate the textual summary of the tool calls.

For \textbf{Script-Style Intent Clustering Knowledge (SIC)}, we first clusters the vector embeddings of instructions using the K-Means algorithm for each scenario with same toolset, with an LLM assigning semantic intent labels to each resulting cluster. For each cluster, a batch-wise extraction and hierarchical induction strategy is employed: task-specific data is processed in batches to circumvent context window limitations, allowing the LLM to summarize intermediate scripts. These scripts, which distill raw tool-calling trajectories into structured JSON Standard Operating Procedures containing conditional logic and step sequences, are further consolidated into a final unified pattern. Finally, these components—comprising intent labels, vector embeddings, and behavioral patterns—are integrated into a searchable procedural memory bank. Following this, for \textbf{Textual-Style Intent Clustering Knowledge (TIC)} the LLM is prompted to generate a natural language textual description for each refined pattern script.

\section{Examples}
\label{sec:appendix_example}
The user's question enhanced with experience knowledge is shown in Example~\ref{prompt:ST}, Example~\ref{prompt:ES}, Example~\ref{prompt:SIC}, Example~\ref{prompt:TIC}. For Script-Style Intent Clustering Knowledge, we provide indentation for readability, but in the actual prompt, there are no line breaks or indentation.

\section{Prompt}
\label{sec:appendix_prompt}
The prompt of LLM-based aggregation of BFCL-V3 is in Prompt~\ref{prompt:aggregation}, and the aggregation prompt of dataset AppWorld is in Prompt~\ref{prompt:aggregation_appworld}. The inference prompt of AppWorld is in Prompt~\ref{prompt:appworld}.
\begin{figure}
    \centering
    \includegraphics[width=0.95\columnwidth]{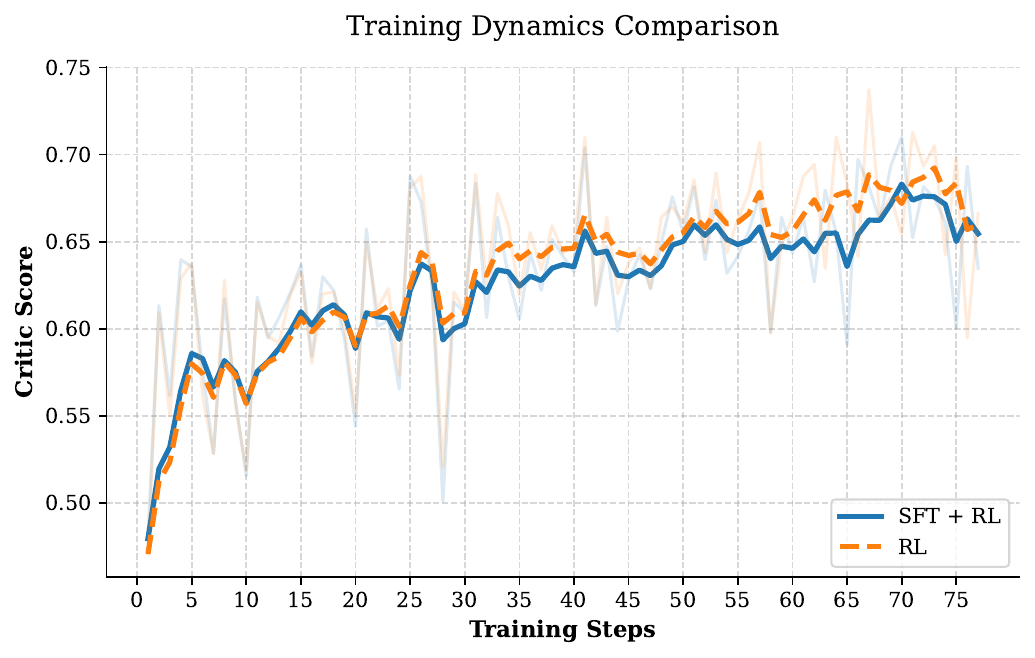}
    \caption{The reward scores of training process of dataset BFCL-V3.}
    \label{fig:curves}
\end{figure}
\begin{figure*}[!ht]
    \centering
    \includegraphics[width=0.95\textwidth]{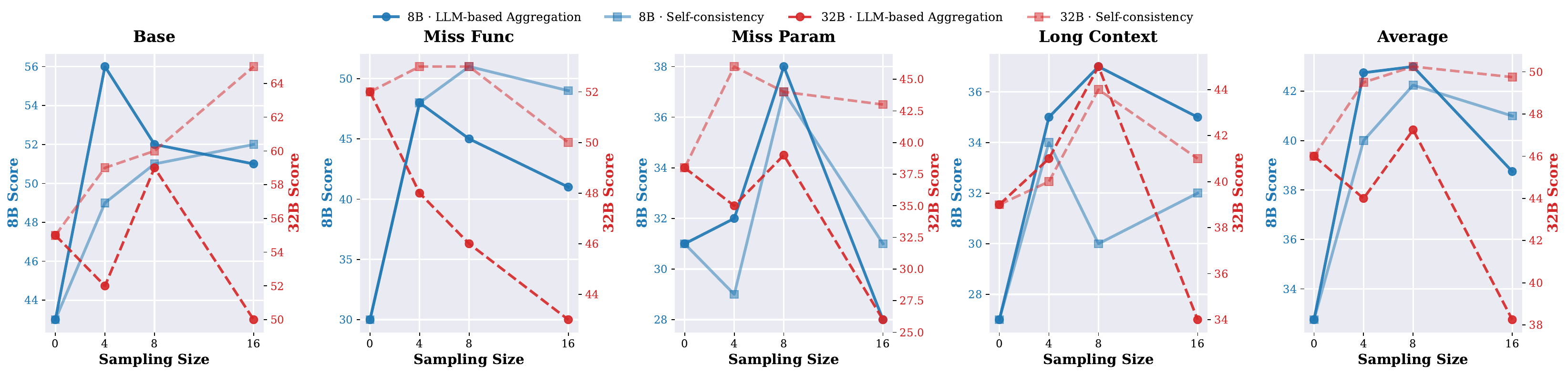}
    \caption{The results of parallel sampling with different sampling size.}
    \label{fig:sampling_results_all}
\end{figure*}

\section{Inference Details}
\label{sec:appendix_inference_details}
\subsection{Retriever Design}
We adopt all-MiniLM-L6-v2 as the retrieval embedding model. 

To optimize retrieval precision across diverse task environments, we design different retriever for datasets. For dataset BFCL, for a given query, the system first encodes the target content into a high-dimensional embedding and restricts the search space to a subset of the knowledge base pre-filtered by the relevant toolset (involved classes). This hard constraint ensures that retrieved experiences are strictly relevant to the required tool operations. Subsequently, candidates undergo similarity thresholding at $p=0.5$, where trajectories with cosine similarity scores below this limit are pruned to maintain high contextual precision. To resolve potential overlaps in multi-category tasks, the framework performs global descending sorting followed by deduplication, ultimately extracting the top  most relevant unique trajectories to serve as experiential guidance for the model. 

For dataset AppWorld, the setting is same as Section~\ref{sec:analysis_setting}. 

\subsection{Inference Details}
We set the temperature to 0 for inference and configured it to 1 for parallel sampling. The parallel sampling size is set to 4, with all experiments conducted on NVIDIA A800 and A100 using Qwen3-8B and Qwen3-32B. We report results from a single run for all experiments; therefore, no error bars or variance statistics are provided.

For dataset BFCL-V3, each testing scenarios containing 200 instances that share identical task descriptions but yield different execution outcomes due to environment-specific dynamics. We select 100 samples from the Base scenario to construct the training set, while the remaining samples are reserved for evaluation. To avoid data leakage, the dataset is split at the sample-ID level, with even-indexed instances assigned to training and odd-indexed instances assigned to testing. All experiential knowledge used for knowledge augmentation is extracted exclusively from the \textit{Base} portion of the training split. 

For dataset AppWorld, we distilled 81 correct examples from the 90 training instances to construct the knowledge base.

For Memp, we follow the ``Proceduralization'' setup from the original paper \cite{DBLP:journals/corr/abs-2508-06433}, that is, we combine trajectories and inductive script as the baseline. Specifically, we use Scenario Trajectory (ST) knowledge and Experience Summary (ES) knowledge as experience, and for the retriever, we adopt the same settings. For ReAct in AppWorld, since the Qwen3 series models output their thought processes by default, we did not ask them to output their thought processes. Instead, we instructed them to output the code directly.

\section{Inference Efficiency}
\label{sec:appendix_inference_efficiency}

To further clarify this trade-off of efficiency and performance introduced by parallel sampling, we have added additional experiments quantifying the computational overhead and latency under different methods in Table~\ref{tab:token_cost_efficiency}. The results show that a limited increase in parallelism yields a disproportionately large improvement in tool-calling accuracy, demonstrating a favorable accuracy–cost trade-off for practical deployment.

\begin{table*}[htbp]
\centering

\begin{tabular}{llccc}
\toprule
Model & Method & Tokens & Time & Token Ratio compared to FC \\
\midrule
\multirow{6}{*}{Qwen3-8B}
& FC & 659,506 & 332 & 1 \\
& Memp & 617,102 & 164 & 0.9357 \\
& KATE & 1,249,096 & 815 & 1.8940 \\
& \quad $\hookrightarrow$ w/o PS & 541,368 & 151 & 0.8209 \\
& \quad $\hookrightarrow$ w/o Exp & 1,671,819 & 1,013 & 2.5350 \\
& \quad $\hookrightarrow$ r PS-Con & 701,391 & 289 & 1.0635 \\
\midrule
\multirow{6}{*}{Qwen3-32B}
& FC & 744,086 & 584 & 1 \\
& Memp & 669,401 & 289 & 0.8996 \\
& KATE & 962,783 & 529 & 1.2939 \\
& \quad $\hookrightarrow$ w/o PS & 571,327 & 346 & 0.7678 \\
& \quad $\hookrightarrow$ w/o Exp & 1,332,647 & 2,351 & 1.7910 \\
& \quad $\hookrightarrow$ r PS-Con & 716,485 & 402 & 0.9629 \\
\bottomrule
\end{tabular}
\caption{Token Consumption and Runtime Comparison.}
\label{tab:token_cost_efficiency}
\end{table*}

We evaluate performance under different methods on 10 data points of Base scenario and report the results. Here, ``r PS-Con'' refers to replacing the LLM-based aggregation with self-consistency, ``w/o PS'' denotes removing parallel sampling, and ``w/o Exp'' indicates removing procedural knowledge. All parallel computations were performed using the maximum multi-threading allowed by vLLM.

The results indicate that, compared to the FC baseline, our method, despite performing four parallel tool calls and final LLM-based aggregation, does not linearly increase inference cost. In fact, the improved accuracy may reduce unnecessary reasoning steps, so the total inference tokens are not multiplied proportionally. For instance, KATE only increases token usage by 1.29× for Qwen3-32B, and using self-consistency nearly matches or even reduces token consumption.

Taken together, while width-based activation introduces additional computation, it aligns with a broader trend in reasoning systems toward test-time scaling, our empirical results show that parallel sampling only adds a small amount of token usage and latency compared to the FC method, while substantially improving accuracy.

\section{Results}
\label{sec:appendix_results}

\begin{table*}[!ht]
    \centering
    \small
    \renewcommand{\arraystretch}{0.9}
    \setlength{\tabcolsep}{5pt}

    \begin{tabular}{l l r r r r r}
        \toprule
        \textbf{Model} & \textbf{Method} & \textbf{Base} & \textbf{Miss F.} & \textbf{Miss P.} & \textbf{Long C.} & \textbf{Average} \\
        \midrule

        \multirow{7}{*}{\textbf{Qwen3-8B}}
        & FC & 45.0 $\pm$ 1.63 & 37.67 $\pm$ 5.56 & 25.67 $\pm$ 4.11 & 29.67 $\pm$ 2.49 & 34.5 $\pm$ 1.34 \\
        & Prompt & 35.0 $\pm$ 3.56 & 34.0 $\pm$ 3.27 & 27.0 $\pm$ 2.16 & 18.67 $\pm$ 1.25 & 28.67 $\pm$ 1.23 \\
        & Memp & 48.0 $\pm$ 3.27 & 28.67 $\pm$ 2.62 & 34.67 $\pm$ 1.89 & 29.0 $\pm$ 2.16 & 35.08 $\pm$ 1.3 \\
        \cmidrule{2-7}

        & \gcell{\textbf{KATE (Ours)}} &
        \gcell{\textbf{58.33 $\pm$ 0.94}} &
        \gcell{40.0 $\pm$ 0.82} &
        \gcell{\textbf{41.67 $\pm$ 0.94}} &
        \gcell{\textbf{40.67 $\pm$ 1.7}} &
        \gcell{\textbf{45.17 $\pm$ 0.92}} \\

        & \gcell{$\quad \hookrightarrow$ w/o PS} &
        \gcell{47.33 $\pm$ 0.94} &
        \gcell{33.33 $\pm$ 1.25} &
        \gcell{33.67 $\pm$ 1.7} &
        \gcell{32.67 $\pm$ 1.25} &
        \gcell{36.75 $\pm$ 1.08} \\

        & \gcell{$\quad \hookrightarrow$ w/o Exp} &
        \gcell{54.33 $\pm$ 1.25} &
        \gcell{\textbf{48.0 $\pm$ 3.27}} &
        \gcell{33.67 $\pm$ 1.25} &
        \gcell{34.33 $\pm$ 1.7} &
        \gcell{42.58 $\pm$ 0.42} \\

        & \gcell{$\quad \hookrightarrow$ r PS-Con.} &
        \gcell{56.67 $\pm$ 0.94} &
        \gcell{35.33 $\pm$ 2.87} &
        \gcell{36.33 $\pm$ 2.49} &
        \gcell{37.67 $\pm$ 1.7} &
        \gcell{41.5 $\pm$ 0.74} \\

        \midrule

        \multirow{7}{*}{\textbf{Qwen3-32B}}
        & FC & 53.67 $\pm$ 0.94 & 50.67 $\pm$ 1.25 & 39.33 $\pm$ 1.25 & 40.0 $\pm$ 1.41 & 45.92 $\pm$ 0.72 \\
        & Prompt & 48.33 $\pm$ 5.31 & 44.33 $\pm$ 3.09 & 36.33 $\pm$ 4.92 & 36.33 $\pm$ 1.7 & 41.33 $\pm$ 2.99 \\
        & Memp & 64.33 $\pm$ 4.19 & 44.67 $\pm$ 1.7 & 46.0 $\pm$ 0.82 & 41.0 $\pm$ 0.82 & 49.0 $\pm$ 0.61 \\
        \cmidrule{2-7}

        & \gcell{\textbf{KATE (Ours)}} &
        \gcell{\textbf{65.33 $\pm$ 2.49}} &
        \gcell{\textbf{52.33 $\pm$ 0.94}} &
        \gcell{44.0 $\pm$ 3.56} &
        \gcell{44.0 $\pm$ 0.82} &
        \gcell{\textbf{51.42 $\pm$ 1.48}} \\

        & \gcell{$\quad \hookrightarrow$ w/o PS} &
        \gcell{61.0 $\pm$ 0.82} &
        \gcell{46.0 $\pm$ 2.45} &
        \gcell{46.0 $\pm$ 1.63} &
        \gcell{42.0 $\pm$ 0.82} &
        \gcell{48.75 $\pm$ 0.35} \\

        & \gcell{$\quad \hookrightarrow$ w/o Exp} &
        \gcell{60.33 $\pm$ 6.55} &
        \gcell{47.67 $\pm$ 0.47} &
        \gcell{39.0 $\pm$ 2.83} &
        \gcell{43.33 $\pm$ 1.7} &
        \gcell{47.58 $\pm$ 2.63} \\

        & \gcell{$\quad \hookrightarrow$ r PS-Con.} &
        \gcell{64.33 $\pm$ 0.94} &
        \gcell{46.33 $\pm$ 2.05} &
        \gcell{\textbf{47.0 $\pm$ 0.82}} &
        \gcell{\textbf{45.33 $\pm$ 0.47}} &
        \gcell{50.75 $\pm$ 0.71} \\

        \bottomrule
    \end{tabular}

    \caption{Performance comparison across models with mean and variance. Green cells denote KATE and its ablations.}
    \label{tab:qwen_results}
\end{table*}

We have conducted additional experiments using multiple random seeds (adding other two running results compared to Table~\ref{tab:bfcl_results}) and now report mean and standard deviation for key results represented in Table~\ref{tab:qwen_results}. 

The findings confirm that:
(1) Width-based activation maintains consistent improvements across seeds.
(2) Moderate parallel sampling improves not only performance but also overall stability compared to greedy decoding.

\begin{table}[!ht]
    \centering
    \small
    \renewcommand{\arraystretch}{0.9}
    \setlength{\tabcolsep}{2pt}
    \begin{tabular}{l r r r r r}
        \toprule
        \textbf{Method} & \textbf{Base} & \textbf{Miss F.} & \textbf{Miss P.} & \textbf{Long C.} & \textbf{Average} \\
        \midrule
        
        FC & 2 & 3 & 1 & 2 & 2.00 \\
        Memp & 18 & 8 & 9 & 8 & 10.75 \\
        \textbf{KATE} & 12 & 15 & 11 & 7 & 11.25 \\
        \quad $\hookrightarrow$ w/o PS & 18 & 14 & 11 & 8 & 12.75 \\
        \quad $\hookrightarrow$ w/o Exp & 7 & 3 & 3 & 4 & 4.25 \\
        \quad $\hookrightarrow$ r PS-Con. & 14 & 15 & 9 & 6 & 11.00 \\
        
        \bottomrule
    \end{tabular}
    \caption{Performance comparison across different methods on Llama3.2-3B-Instruct.}
    \label{tab:result_llama}
\end{table}

As shown in Table~\ref{tab:result_llama}, the results confirm the effectiveness of our method, demonstrating that KATE performance beyond Qwen3 model. Here, ``r PS-Con'' refers to replacing the LLM-based aggregation with self-consistency, ``w/o PS'' denotes removing parallel sampling, and ``w/o Exp'' indicates removing procedural knowledge.

We set the temperature to 0.02, as we found that using a higher temperature for this model may lead to incorrect tool-calling formats in the generated outputs.

The experimental results show that both parallel sampling (w/o Exp) and procedural knowledge (w/o PS) significantly improve model performance. KATE consistently outperforms the baseline. Although its improvement over the procedural-knowledge-only variant (w/o Exp) is not substantial, this may be because the model is already approaching its performance ceiling, leaving limited room for further gains.

\section{Training Details}
\label{sec:appendix_training_details}
We augment the training data with experiential knowledge and decompose multi-turn tool-calling sequences into individual turns across different rounds. The resulting dataset is split 1:1 for Supervised Fine-Tuning (SFT) and Reinforcement Learning (RL), with samples exceeding a text length of 8192 tokens removed. For the RL process, we extract the tool-calling outputs and verify their correctness using a matching-based evaluation.

We fine-tune the model using both Supervised Fine-Tuning (SFT) and Reinforcement Learning (RL), adopting LoRA-based parameter-efficient tuning across all stages. For SFT, we set the learning rate to 3e-5, train for 3 epochs, and use a LoRA rank of 32 with LoRA alpha set to 16. For RL, we use GRPO method. We use a training batch size of 128, a maximum prompt length of 8192, and a maximum response length of 2048. The learning rate is also set to 3e-5, with 8 sampled trajectories per prompt and 7 training epochs, while maintaining a LoRA rank of 32.

We use all models and datasets in compliance with their licenses.
\begin{algorithm}[t]
\caption{Multi-turn Parallel Action Sampling with Aggregation} 
\label{alg:parallel_multi_turn_tool_use}
\begin{algorithmic} 
\Require 
Initial dialogue history $\mathcal{H}_0$; 
tool set $\mathcal{T}$; 
system prompt $S$; 
{\color{darkred}parallel sample size $N$}; 
{\color{darkred}aggregation function $\mathcal{A}(\cdot)$}; 
maximum steps $T$

\State $\mathcal{H}_t \leftarrow \mathcal{H}_0$, $t \leftarrow 0$;

\While{$t < T$}
    \State \textcolor{darkred}{$\textit{actions} \leftarrow \emptyset$};
    \ForAll{$i = 1, \dots, N$ \textbf{in parallel}}
        \State \textcolor{darkred}{$o^{(i)}_{t+1} \sim P(o_{t+1} \mid \mathcal{T}, S, \mathcal{H}_t)$};
        \State \textcolor{darkred}{$\textit{actions} \leftarrow \textit{actions} \cup \{o^{(i)}_{t+1}\}$};
    \EndFor

    \If{All samples in $\textit{actions}$ are identical}
        \State $o_{t+1} \leftarrow o^{(1)}_{t+1}$;
    \Else
        \State \textcolor{darkred}{$o_{t+1} \leftarrow \mathcal{A}(\textit{actions} \mid S, \mathcal{H}_t)$};
    \EndIf

    \If{$o_{t+1}$ is a tool invocation $c_{t+1}$}
        \State Execute $c_{t+1}$ and observe environment reward $r^{\text{env}}_{t+1}$;
        \State $\mathcal{H}_{t+1} \leftarrow \mathcal{H}_t \cup \{c_{t+1}, r^{\text{env}}_{t+1}\}$;
    \Else
        \If{No further user query}
            \State \Return $o_{t+1}$;
        \EndIf
        \State Observe user reply $r^{\text{user}}_{t+1}$;
        \State $\mathcal{H}_{t+1} \leftarrow \mathcal{H}_t \cup \{o_{t+1}, r^{\text{user}}_{t+1}\}$;
    \EndIf
    \State $t \leftarrow t + 1$;
\EndWhile
\end{algorithmic}
\end{algorithm}

\begin{algorithm}[t]
\caption{KATE}
\label{alg:stepwise_tool_use}
\begin{algorithmic} 
\Require 
Initial history $\mathcal{H}_0$; tool ses $\mathcal{T}$; system prompt $S$; 
{\color{darkred}Parallel size $N$}; 
{\color{darkred}Similarity threshold $p$}; 
{\color{darkred}Aggregation function $\mathcal{A}(\cdot)$};
maximum steps $T$.

\State Initialize $\mathcal{H}_0^{\text{re}} \leftarrow \mathcal{H}_0$, $t \leftarrow 0$;

\While{$t < T$}

    \State Observe user query $r_t = r_t^{\text{user}}$;
    
    \State $\mathcal{H}_t^{\text{re}} \leftarrow \mathcal{H}_t^{\text{re}} \cup \mathcal{R}(r_t^{\text{user}})$;

    \State \textcolor{darkred}{Initialize $\textit{actions} \leftarrow \emptyset$;}
    \ForAll{$i = 1, \dots, N$ \textbf{in parallel}}
        \State \textcolor{darkred}{$o^{(i)}_{t+1} \sim P(o_{t+1} \mid \mathcal{T}, S, \mathcal{H}_t^{\text{re}}$);}
        \State \textcolor{darkred}{$\textit{actions} \leftarrow \textit{actions} \cup \{o^{(i)}_{t+1}\}$;}
    \EndFor

    \If{All samples in $\textit{actions}$ are identical}
        \State $o_{t+1} \leftarrow o^{(1)}_{t+1}$;
    \Else
        \State \textcolor{darkred}{$o_{t+1} \leftarrow \mathcal{A}(\textit{actions} \mid S, \mathcal{H}_t^{\text{re}})$};
    \EndIf

    \If{$o_{t+1}$ is a tool invocation $c_{t+1}$}
        \State Execute $c_{t+1}$ and observe environment feedback $r^{\text{env}}_{t+1}$;
        \State $\mathcal{H}_{t+1}^{\text{re}} \leftarrow \mathcal{H}_t^{\text{re}} \cup \{c_{t+1}, r^{\text{env}}_{t+1}\}$;
        
    \Else
        \If{No further user query}
            \State \Return $o_{t+1}$;
        \EndIf
        \State Observe user reply $r_{t+1}^{\text{user}}$
        \State $\mathcal{H}_{t+1}^{\text{re}} \leftarrow \mathcal{H}_{t}^{\textbf{re}} \cup \{o_{t+1}, r_{t+1}^{\text{user}}\}$;

    \EndIf
    \State $t \leftarrow t + 1$;
\EndWhile
\end{algorithmic}
\end{algorithm}
\twocolumn

\newpage

\onecolumn

\begin{tcolorbox}[width=\textwidth, breakable]
{
\renewcommand{\lstlistingname}{Prompt}
\begin{lstlisting}[
    frame=none,
    basicstyle=\ttfamily\footnotesize,
    breaklines=true,
    breakatwhitespace=true,
    breakindent=0pt,
    columns=fullflexible,
    caption={Aggregation Prompt used in KATE on dataset BFCL-V3},
    captionpos=b,
    label={prompt:aggregation}
]
You are a tool calling agent. Based on the conversation history, available tools, and candidate tool calls provided.
Your task is to evaluate multiple candidate tool calls generated for the user's questions and assistant responses, analyze their correctness, and produce a single **optimal plan** along with a **validated tool call**.

---

### Inputs
- Candidate tool calls: {candidate_plans}  

**Return Format**  
   Return a JSON object with the following structure:

```json
{{
  "optimal_plan": "<Explain The optimal plan and tool calls to execute next (You don't need to explain why you choose this approach, but rather explain why you are executing this tool_call.)>",
  "optimal_tool_call": {{
    "name": "<tool name>",
    "parameters": {{}}
  }}
}}
Only one tool call is allowed in the optimal_tool_call.
If no tool call is needed, set "optimal_tool_call": {{"name": "response_to_user", "parameters": {{"content": "The response to the user"}}}}.
\end{lstlisting}
}
\end{tcolorbox}


\begin{tcolorbox}[width=\textwidth, breakable]
{
\renewcommand{\lstlistingname}{Prompt}
\begin{lstlisting}[
    frame=none,
    basicstyle=\ttfamily\footnotesize,
    breaklines=true,
    breakatwhitespace=true,
    breakindent=0pt,
    columns=fullflexible,
    caption={Aggregation Prompt used in KATE on dataset AppWorld},
    captionpos=b,
    label={prompt:aggregation_appworld}
]
You are a highly skilled assistant tasked with generating Python code in a **step-by-step** manner. Your goal is to progressively generate the correct code based on the conversation history and multiple candidate solutions. After each step, you should assess the results of the generated code and, if needed, iterate to make improvements. You should not attempt to generate all the code at once. Instead, generate a small portion of the code at a time, test it, and refine it based on the feedback received. If previous code attempts were incorrect, reassess the logic and generate the next step of code accordingly.

---

### Input Information:

- **User's Question**:
  {user_question}

- **Interaction History Between Assistant and the Environment**:
  {interaction_history}

- **Candidate Code Options**:
  {candidate_code}

---

### Instructions:

1. **Understand the Problem**: Thoroughly review the user's question and the interaction history to grasp the context.
2. **Break Down the Task**: Start by identifying the most critical portion of the code that needs to be addressed first. Focus on one current step at a time.
3. **Generate a Small Portion of Code**: Produce only the part of the code needed to address the first step of the task. Do not try to complete the entire solution in one go.
4. **Iterate Based on Feedback**: If previous steps have errors or issues, do not just fix them all at once. Focus on understanding the specific issue and generating the next part of the solution, keeping previous code intact as much as possible.
5. **Iterative Refinement**: With each step, you should refine your approach based on what has been known so far, gradually moving towards a complete solution.

---

### Output Format:

```python
# The code for the current step in the task.
Add your code here
```
\end{lstlisting}
}
\end{tcolorbox}


\begin{tcolorbox}[width=\textwidth, breakable]
{
\renewcommand{\lstlistingname}{Prompt}
\begin{lstlisting}[
    frame=none,
    basicstyle=\ttfamily\footnotesize,
    breaklines=true,
    breakatwhitespace=true,
    breakindent=0pt,
    columns=fullflexible,
    caption={Intent prompt in Depth-based Prompt-Hint Activation},
    captionpos=b,
    label={prompt:intent}
]
Before invoking any tools, clearly identify the user's *current intent* based on the conversation history and the latest user message.

There are some suggestions for your reasoning:
1. Identify the user's current intent based on the conversation history and the latest user message.
2. Break down this intent into clear, actionable subtasks or goals.
3. Determine which tools (if any) are needed for each subtask, and specify their expected inputs and outputs. Your reasoning should focus on *clarity* (what the user wants), *structure* (how to achieve it), and *efficiency* (which tool or reasoning step should come next).

You do not have to fully adhere to the above suggestions. But you need to analyze the relevant points in the conversation history about the intent requirements in the thinking process.
\end{lstlisting}
}
\end{tcolorbox}



\begin{tcolorbox}[width=\textwidth, breakable]
{
\renewcommand{\lstlistingname}{Prompt}
\begin{lstlisting}[
    frame=none,
    basicstyle=\ttfamily\footnotesize,
    breaklines=true,
    breakatwhitespace=true,
    breakindent=0pt,
    columns=fullflexible,
    caption={Reflection prompt in Depth-based Prompt-Hint Activation},
    captionpos=b,
    label={prompt:reflection}
]
Before invoking new tools, review the history of tool calls and their outcomes.

There are some suggestions for your reasoning:
1. Determine whether the previous tool calls were correct, sufficient, or complete. If a tool call failed or produced suboptimal results due to insufficient or missing parameters or functions, reflect on what information was lacking, how it could be inferred or obtained.
2. If issues exist (e.g., wrong parameters, missing calls, failed execution), explain briefly why they occurred.
3. Analyze future multi-step tool calls during the analysis process, rather than just focusing on the next step.

You do not have to fully adhere to the above suggestions. But you need to analyze the relevant points in the conversation history about the correctness and necessary of previous tool call in the thinking process.
\end{lstlisting}
}
\end{tcolorbox}



\begin{tcolorbox}[width=\textwidth, breakable]
{
\renewcommand{\lstlistingname}{Prompt}
\begin{lstlisting}[
    frame=none,
    basicstyle=\ttfamily\footnotesize,
    breaklines=true,
    breakatwhitespace=true,
    breakindent=0pt,
    columns=fullflexible,
    caption={State prompt in Depth-based Prompt-Hint Activation},
    captionpos=b,
    label={prompt:state}
]
Before invoking any tools, carefully identify the *environment* and the *state* required to answer the user's question, because these may influence both the tool selection and the parameters for tool calls.

There are some suggestions for your reasoning:
1. Analyze the user's question to detect any implicit state dependencies (e.g., user login status, file existence, context variables).
2. Determine what specific states must be confirmed before continuing.
3. If verification is required, decide which tools should be invoked to confirm those states. If no state verification is needed, proceed with reasoning toward tool selection or response generation.

You do not have to fully adhere to the above suggestions. But you need to analyze the relevant points in the conversation history about the state requirements in the thinking process.
\end{lstlisting}
}
\end{tcolorbox}



\begin{tcolorbox}[width=\textwidth, breakable]
{
\renewcommand{\lstlistingname}{Prompt}
\begin{lstlisting}[
    frame=none,
    basicstyle=\ttfamily\footnotesize,
    breaklines=true,
    breakatwhitespace=true,
    breakindent=0pt,
    columns=fullflexible,
    caption={Prompt of AppWorld},
    captionpos=b,
    label={prompt:appworld}
]
USER:
I am your supervisor and you are a super intelligent AI Assistant whose job is to achieve my day-to-day tasks completely autonomously.

To do this, you will need to interact with app/s (e.g., spotify, venmo, etc) using their associated APIs on my behalf. For this you will undertake a *multi-step conversation* using a python REPL environment. That is, you will write the python code and the environment will execute it and show you the result, based on which, you will write python code for the next step and so on, until you've achieved the goal. This environment will let you interact with app/s using their associated APIs on my behalf.

Here are three key APIs that you need to know to get more information

# To get a list of apps that are available to you.
print(apis.api_docs.show_app_descriptions())

# To get the list of apis under any app listed above, e.g. supervisor
print(apis.api_docs.show_api_descriptions(app_name='supervisor'))

# To get the specification of a particular api, e.g. supervisor app's show_account_passwords
print(apis.api_docs.show_api_doc(app_name='supervisor', api_name='show_account_passwords'))

Each code execution will produce an output that you can use in subsequent calls. Using these APIs, you can now generate code, that the environment will execute, to solve the task.

For example, consider the task:

My name is: {{ supervisor.first_name }} {{ supervisor.last_name }}. My personal email is {{ supervisor.email }} and phone number is {{ supervisor.phone_number }}.

Task:

What is the password for my Spotify account?

ASSISTANT:
# Okay. Lets first find which apps are available to get the password by looking at the app descriptions.
print(apis.api_docs.show_app_descriptions())

USER:
[
  {
    "name": "api_docs",
    "description": "An app to search and explore API documentation."
  },
  {
    "name": "supervisor",
    "description": "An app to access supervisor's personal information, account credentials, addresses, payment cards, and manage the assigned task."
  },
  ...
  {
    "name": "spotify",
    "description": "A music streaming app to stream songs and manage song, album and playlist libraries."
  },
  {
    "name": "venmo",
    "description": "A social payment app to send, receive and request money to and from others."
  },
  ...
]


ASSISTANT:
# Looks like the supervisor app could help me with that. Lets see what apis are available under this app.
print(apis.api_docs.show_api_descriptions(app_name='supervisor'))


USER:
[
  ...
  "show_account_passwords : Show your supervisor's account passwords."
  ...
]


ASSISTANT:
# I can use `show_account_passwords` to get the passwords. Let me see its detailed specification to understand its arguments and output structure.
print(apis.api_docs.show_api_doc(app_name='supervisor', api_name='show_account_passwords'))

USER:
{
  'app_name': 'supervisor',
  'api_name': 'show_account_passwords',
  'path': '/account_passwords',
  'method': 'GET',
  'description': "Show your supervisor's app account passwords.",
  'parameters': [],
  'response_schemas': {
    'success': [{'account_name': 'string', 'password': 'string'}],
    'failure': {'message': 'string'}
  }
}

ASSISTANT:
# Okay, it requires no arguments. So I can just call it directly.
print(apis.supervisor.show_account_passwords())

USER:
[
  {
    "account_name": "spotify",
    "password": "dummy_spotify_pass"
  },
  {
    "account_name": "file_system",
    "password": "dummy_fs_pass"
  },
  ...
]


ASSISTANT:
# So the Spotify password is an entry in the `passwords` list with the account_name=spotify.
spotify_password = [account_password["account_name"] == "spotify" for account_password in passwords][0]["password"]
print(spotify_password)


USER:
dummy_spotify_pass

ASSISTANT:
# When the task is completed, I need to call apis.supervisor.complete_task(). If there is an answer, I need to pass it as an argument `answer`. I will pass the spotify_password as an answer.
apis.supervisor.complete_task(answer=spotify_password)


USER:
Marked the active task complete.


----------------------------------------------

USER:
**Key instructions and disclaimers**:

1. The email addresses, access tokens and variables (e.g. spotify_password) in the example above were only for demonstration. Obtain the correct information by calling relevant APIs yourself.
2. Only generate valid code blocks, i.e., do not put them in ```...``` or add any extra formatting. Any thoughts should be put as code comments.
3. You can use the variables from the previous code blocks in the subsequent code blocks.
4. Write small chunks of code and only one chunk of code in every step. Make sure everything is working correctly before making any irreversible change.
5. The provided Python environment has access to its standard library. But modules and functions that have a risk of affecting the underlying OS, file system or process are disabled. You will get an error if do call them.
6. Any reference to a file system in the task instructions means the file system *app*, operable via given APIs, and not the actual file system the code is running on. So do not write code making calls to os-level modules and functions.
7. To interact with apps, only use the provided APIs, and not the corresponding Python packages. E.g., do NOT use `spotipy` for Spotify. Remember, the environment only has the standard library.
8. The provided API documentation has both the input arguments and the output JSON schemas. All calls to APIs and parsing its outputs must be as per this documentation.
9. For APIs that return results in "pages", make sure to consider all pages.
10. To obtain current date or time, use Python functions like `datetime.now()` or obtain it from the phone app. Do not rely on your existing knowledge of what the current date or time is.
11. For all temporal requests, use proper time boundaries, e.g., if I ask for something that happened yesterday, make sure to consider the time between 00:00:00 and 23:59:59. All requests are concerning a single, default (no) time zone.
12. Any reference to my friends, family or any other person or relation refers to the people in my phone's contacts list.
13. All my personal information, and information about my app account credentials, physical addresses and owned payment cards are stored in the "supervisor" app. You can access them via the APIs provided by the supervisor app.
14. Once you have completed the task, call `apis.supervisor.complete_task()`. If the task asks for some information, return it as the answer argument, i.e. call `apis.supervisor.complete_task(answer=<answer>)`. For tasks that do not require an answer, just skip the answer argument or pass it as None.
15. The answers, when given, should be just entity or number, not full sentences, e.g., `answer=10` for "How many songs are in the Spotify queue?". When an answer is a number, it should be in numbers, not in words, e.g., "10" and not "ten".
16. You can also pass `status="fail"` in the complete_task API if you are sure you cannot solve it and want to exit.
17. You must make all decisions completely autonomously and not ask for any clarifications or confirmations from me or anyone else.

### Instructions:

1. **Understand the Problem**: Thoroughly review the user's question and the interaction history to grasp the context.
2. **Break Down the Task**: Start by identifying the most critical portion of the code that needs to be addressed first. Focus on one current step at a time.
3. **Generate a Small Portion of Code**: Produce only the part of the code needed to address the first step of the task. Do not try to complete the entire solution in one go.
4. **Iterate Based on Feedback**: If previous steps have errors or issues, do not just fix them all at once. Focus on understanding the specific issue and generating the next part of the solution, keeping previous code intact as much as possible.
5. **Iterative Refinement**: With each step, you should refine your approach based on what has been known so far, gradually moving towards a complete solution.

And you need to call apis.api_docs.show_app_descriptions(), apis.api_docs.show_api_descriptions(app_name='<app>') and apis.api_docs.show_api_doc(app_name='<app>', api_name='<app_name>') before utilizing <app> and <app_name> others at first time. And you need to print the result of each API call.

You don't need to generate the entire code at once. You can generate the code step by step and execute it.

USER:
Using these APIs, now generate code to solve the actual task:

My name is: {{ supervisor.first_name }} {{ supervisor.last_name }}. My personal email is {{ supervisor.email }} and phone number is {{ supervisor.phone_number }}.

Task:

{{ instruction }}
\end{lstlisting}
}
\end{tcolorbox}



\begin{tcolorbox}[width=\textwidth, breakable]
{
\renewcommand{\lstlistingname}{Prompt}
\begin{lstlisting}[
    frame=none,
    basicstyle=\ttfamily\footnotesize,
    breaklines=true,
    breakatwhitespace=true,
    breakindent=0pt,
    columns=fullflexible,
    caption={Example of Scenario Trajectory Knowledge (ST)},
    captionpos=b,
    label={prompt:ST}
]
Original Question:
How much would a flight from SF to LA even cost? It's probably cheap. That's it, let me just do it. I need to arrange a business class flight for Robert Trenton from San Francisco to Los Angeles on November 25th 2024. The reservation should be made using his travel card with id card_3487 and access code 1293. Following the booking, I have to ensure an invoice is issued to verify the charges.

Enhanced Question:
How much would a flight from SF to LA even cost? It's probably cheap. That's it, let me just do it. I need to arrange a business class flight for Robert Trenton from San Francisco to Los Angeles on November 25th 2024. The reservation should be made using his travel card with id card_3487 and access code 1293. Following the booking, I have to ensure an invoice is issued to verify the charges.

Before answering the user's question above, please first review the following related experiences:

### Example 1
**Question:** I'm planning to fly from San Francisco to Los Angeles on October 15, 2024. Could you assist in securing a first-class seat using my travel card with id 'travel_card_12345'? Everything you need—access token (abc123xyz456), traveler details—are at the ready. Just make sure that I can afford it because I only have 6000 dollars to spend for this flight.

**Correct Tool Calling Trajectory for Reference:**
- get_flight_cost(travel_from='SFO', travel_to='LAX', travel_date='2024-10-15', travel_class='first')
- book_flight(access_token='abc123xyz456', card_id='travel_card_12345', travel_date='2024-10-15', travel_from='SFO', travel_to='LAX', travel_class='first')

### Example 2
**Question:** Now, with the newly set budget and using card with id 1432 out of my available cards, I'd like to book that business-class flight from Rivermist to Los Angeles on August 15, 2024, utilizing my access token ABCDE12345. You already know the travel cost!

**Correct Tool Calling Trajectory for Reference:**
- book_flight(access_token='ABCDE12345', card_id='1432', travel_date='2024-08-15', travel_from='RMS', travel_to='LAX', travel_class='business')

### Example 3
**Question:** I'm planning to jet off and stumbled upon a nifty flight from San Francisco to Los Angeles. Lowkey, my budget isn't too tight, but just make sure overall cost for the flight is below $10,000? You should know my travel class is 'first' and travel date is '2024-11-15'. Oh, and I've got my card with id 'card123' and account token 'access_token_abc123' all good to go.

**Correct Tool Calling Trajectory for Reference:**
- get_flight_cost(travel_from='SFO',travel_to='LAX',travel_date='2024-11-15',travel_class='first')
- book_flight(access_token='access_token_abc123', card_id='card123', travel_date='2024-11-15', travel_from='SFO', travel_to='LAX', travel_class='first')


**Note**: You are not required to reference the information or examples above if they are not directly relevant to the current user question. Analyze the problem carefully, decide whether the retrieved information is useful, and always apply reasoning before making any tool calls.
Your actions must be based on the information given by the current user. You can not make up data, nor can you refer to examples that will cause you to act beyond the current information.
You need to determine the difference between your question and the question in retrieval examples.
Attention the user question at current turn is: 
How much would a flight from SF to LA even cost? It's probably cheap. That's it, let me just do it. I need to arrange a business class flight for Robert Trenton from San Francisco to Los Angeles on November 25th 2024. The reservation should be made using his travel card with id card_3487 and access code 1293. Following the booking, I have to ensure an invoice is issued to verify the charges.
\end{lstlisting}
}
\end{tcolorbox}



\begin{tcolorbox}[width=\textwidth, breakable]
{
\renewcommand{\lstlistingname}{Prompt}
\begin{lstlisting}[
    frame=none,
    basicstyle=\ttfamily\footnotesize,
    breaklines=true,
    breakatwhitespace=true,
    breakindent=0pt,
    columns=fullflexible,
    caption={Example of Experience Summary Knowledge (ES)},
    captionpos=b,
    label={prompt:ES}
]
Original Question:
How much would a flight from SF to LA even cost? It's probably cheap. That's it, let me just do it. I need to arrange a business class flight for Robert Trenton from San Francisco to Los Angeles on November 25th 2024. The reservation should be made using his travel card with id card_3487 and access code 1293. Following the booking, I have to ensure an invoice is issued to verify the charges.

Enhanced Question:
How much would a flight from SF to LA even cost? It's probably cheap. That's it, let me just do it. I need to arrange a business class flight for Robert Trenton from San Francisco to Los Angeles on November 25th 2024. The reservation should be made using his travel card with id card_3487 and access code 1293. Following the booking, I have to ensure an invoice is issued to verify the charges.

Before answering the user's question above, please first review the following related experiences:

### Example 1
**Question:** I'm planning to fly from San Francisco to Los Angeles on October 15, 2024. Could you assist in securing a first-class seat using my travel card with id 'travel_card_12345'? Everything you need—access token (abc123xyz456), traveler details—are at the ready. Just make sure that I can afford it because I only have 6000 dollars to spend for this flight.

**Analysis & Advice:**
To solve the user's request accurately, the task involved extracting essential information such as the travel route (San Francisco to Los Angeles), date (October 15, 2024), and class (first) from the question. The user's budget constraint of $6000 was identified, necessitating a check on the flight cost before booking. The correct tools were selected: `get_flight_cost` to verify affordability and `book_flight` to confirm booking details using the provided access token and travel card ID. This structured approach ensured that parameters aligned with user needs. For future tasks, apply a systematic extraction of constraints, verify budget compatibility before booking, and ensure tool parameters match the user's context to maintain accuracy and reliability.

### Example 2
**Question:** Now, with the newly set budget and using card with id 1432 out of my available cards, I'd like to book that business-class flight from Rivermist to Los Angeles on August 15, 2024, utilizing my access token ABCDE12345. You already know the travel cost!

**Analysis & Advice:**
In the current turn, the problem was accurately interpreted by identifying the user's request to book a flight using specific constraints: budget, card ID, and access token. The task required extracting these parameters from the user's message and using them to configure the `book_flight` tool call. The constraints were validated against the user's context from previous turns, ensuring the travel cost and airport code were already known. This task's success relied on precise parameter extraction and validation. For future tasks, consistently extract parameters from user input and context, validate them against known data, and select tools that align with task requirements. Ensure all parameters are mapped correctly and verified for consistency with previous interactions, and use error handling to manage potential inconsistencies.

### Example 3
**Question:** I'm planning to jet off and stumbled upon a nifty flight from San Francisco to Los Angeles. Lowkey, my budget isn't too tight, but just make sure overall cost for the flight is below $10,000? You should know my travel class is 'first' and travel date is '2024-11-15'. Oh, and I've got my card with id 'card123' and account token 'access_token_abc123' all good to go.

**Analysis & Advice:**
In the current task, the agent successfully interpreted the user's request by identifying key constraints: departure ('SFO'), destination ('LAX'), travel date ('2024-11-15'), travel class ('first'), and budget (below $10,000). The correct tools were selected: `get_flight_cost` to check the cost against the budget, and `book_flight` to finalize the booking using provided credentials ('card123', 'access_token_abc123'). This demonstrates effective constraint extraction and parameter mapping. For future tasks, ensure clear identification of constraints and credentials from user input, verify tool capabilities (e.g., cost retrieval before booking), and confirm parameters meet user criteria before proceeding with actions. This structured approach improves accuracy and reliability in handling similar tasks.


**Note**: You are not required to reference the information or examples above if they are not directly relevant to the current user question. Analyze the problem carefully, decide whether the retrieved information is useful, and always apply reasoning before making any tool calls.
Your actions must be based on the information given by the current user. You can not make up data, nor can you refer to examples that will cause you to act beyond the current information.
You need to determine the difference between your question and the question in retrieval examples.
Attention the user question at current turn is: 
How much would a flight from SF to LA even cost? It's probably cheap. That's it, let me just do it. I need to arrange a business class flight for Robert Trenton from San Francisco to Los Angeles on November 25th 2024. The reservation should be made using his travel card with id card_3487 and access code 1293. Following the booking, I have to ensure an invoice is issued to verify the charges.
\end{lstlisting}
}
\end{tcolorbox}



\begin{tcolorbox}[width=\textwidth, breakable]
{
\renewcommand{\lstlistingname}{Prompt}
\begin{lstlisting}[
    frame=none,
    basicstyle=\ttfamily\footnotesize,
    breaklines=true,
    breakatwhitespace=true,
    breakindent=0pt,
    columns=fullflexible,
    caption={Example of Script-Style Intent Clustering Knowledge (SIC)},
    captionpos=b,
    label={prompt:SIC}
]
Original Question:
I've just secured all doors and engaged the parking brake in my vehicle, please start the engine with the ignition on START mode.

Enhanced Question:
I've just secured all doors and engaged the parking brake in my vehicle, please start the engine with the ignition on START mode.

Before answering the user's question above, please first review the following related experiences:

The user's intent is The user intends to start the vehicle's engine by engaging the ignition in START mode after ensuring all doors are secured and the parking brake is applied.
There are some behavior pattern for you to reference:
**Pattern of The common intent is to ensure car doors are locked and the engine is started for a safe and prepared journey.**: [
  {
    "note": "Preconditions and safety: ensure the user is authorized to control the vehicle, the vehicle is in Park/Neutral, and surroundings are safe."
  },
  {
    "call": "displayCarStatus",
    "args": {
      "option": "doors"
    },
    "save_as": "doors_status"
  },
  {
    "condition": "<unlock_requested> == True",
    "then": [
      {
        "call": "lockDoors",
        "args": {
          "unlock": true,
          "door": [
            "driver",
            "passenger",
            "rear_left",
            "rear_right"
          ]
        },
        "save_as": "unlock_result"
      },
      {
        "condition": "unlock_result.lockStatus == 'unlocked'",
        "then": [
          {
            "note": "Doors successfully unlocked as requested."
          }
        ],
        "else": [
          {
            "action": "fail",
            "reason": "Failed to unlock all requested doors."
          }
        ]
      }
    ],
    "else": [
      {
        "condition": "doors_status.status.doorStatus contains any 'unlocked'",
        "then": [
          {
            "call": "lockDoors",
            "args": {
              "unlock": false,
              "door": [
                "driver",
                "passenger",
                "rear_left",
                "rear_right"
              ]
            },
            "save_as": "lock_result"
          },
          {
            "condition": "lock_result.lockStatus == 'locked' and lock_result.remainingUnlockedDoors == 0",
            "then": [
              {
                "note": "All doors locked successfully."
              }
            ],
            "else": [
              {
                "call": "displayCarStatus",
                "args": {
                  "option": "doors"
                },
                "save_as": "doors_status_after"
              },
              {
                "condition": "doors_status_after.status.doorStatus contains any 'unlocked'",
                "then": [
                  {
                    "action": "fail",
                    "reason": "Unable to lock all doors after retry."
                  }
                ]
              }
            ]
          }
        ],
        "else": [
          {
            "note": "All doors were already locked."
          }
        ]
      }
    ]
  },
  {
    "condition": "<ensure_parking_brake_engaged> == True",
    "then": [
      {
        "call": "activateParkingBrake",
        "args": {
          "mode": "engage"
        },
        "save_as": "parking_brake_status"
      }
    ]
  },
  {
    "call": "pressBrakePedal",
    "args": {
      "pedalPosition": 1.0
    },
    "save_as": "brake_press"
  },
  {
    "condition": "brake_press.brakePedalStatus == 'pressed'",
    "then": [
      {
        "call": "startEngine",
        "args": {
          "ignitionMode": "START"
        },
        "save_as": "engine_start"
      }
    ],
    "else": [
      {
        "action": "fail",
        "reason": "Brake pedal not pressed; cannot start engine."
      }
    ]
  },
  {
    "condition": "engine_start.engineState == 'running'",
    "then": [
      {
        "note": "Engine started successfully."
      },
      {
        "call": "displayCarStatus",
        "args": {
          "option": "engine"
        },
        "save_as": "engine_status"
      },
      {
        "call": "displayCarStatus",
        "args": {
          "option": "fuel"
        },
        "save_as": "fuel_status"
      },
      {
        "call": "displayCarStatus",
        "args": {
          "option": "battery"
        },
        "save_as": "battery_status"
      },
      {
        "condition": "<need_climate_status> == True",
        "then": [
          {
            "call": "displayCarStatus",
            "args": {
              "option": "climate"
            },
            "save_as": "climate_status"
          }
        ]
      },
      {
        "condition": "<need_headlights> == True",
        "then": [
          {
            "call": "setHeadlights",
            "args": {
              "mode": "<headlight_mode_on_or_auto>"
            },
            "save_as": "headlights_status"
          }
        ]
      },
      {
        "condition": "<check_tire_pressure> == True",
        "then": [
          {
            "call": "check_tire_pressure",
            "args": {},
            "save_as": "tire_status"
          },
          {
            "condition": "tire_status.tirePressure.healthy_tire_pressure == False",
            "then": [
              {
                "call": "find_nearest_tire_shop",
                "args": {},
                "save_as": "tire_shop"
              }
            ]
          }
        ]
      },
      {
        "condition": "<ready_to_drive> == True",
        "then": [
          {
            "call": "activateParkingBrake",
            "args": {
              "mode": "release"
            },
            "save_as": "parking_brake_release"
          },
          {
            "call": "releaseBrakePedal",
            "args": {},
            "save_as": "brake_release"
          }
        ],
        "else": [
          {
            "note": "Remain stationary: keep parking brake engaged and foot brake as needed."
          }
        ]
      }
    ],
    "else": [
      {
        "note": "Engine failed to start; proceed with diagnostics."
      },
      {
        "call": "displayCarStatus",
        "args": {
          "option": "fuel"
        },
        "save_as": "fuel_status_on_fail"
      },
      {
        "call": "displayCarStatus",
        "args": {
          "option": "battery"
        },
        "save_as": "battery_status_on_fail"
      },
      {
        "condition": "fuel_status_on_fail.status.fuelLevel <= <fuel_min_threshold_gal>",
        "then": [
          {
            "call": "fillFuelTank",
            "args": {
              "fuelAmount": "<fuel_amount_gal>"
            },
            "save_as": "refuel_result"
          }
        ]
      },
      {
        "action": "fail",
        "reason": "Engine did not start; check battery voltage and fuel level, then retry."
      }
    ]
  },
  {
    "note": "Optional: If user explicitly requests door unlocking (e.g., for access), run the unlock branch above before any start steps. If low-light is expected or requested, set headlights to 'auto' or 'on'. Always confirm critical states (doors, engine, fuel, battery) after actions."
  }
]
**Note**: You are not required to reference the information or examples above if they are not directly relevant to the current user question. Analyze the problem carefully, decide whether the retrieved information is useful, and always apply reasoning before making any tool calls.
Your actions must be based on the information given by the current user. You can not make up data, nor can you refer to examples that will cause you to act beyond the current information.
You need to determine the difference between your question and the question in retrieval examples.
Attention the user question at current turn is: 
I've just secured all doors and engaged the parking brake in my vehicle, please start the engine with the ignition on START mode.
\end{lstlisting}
}
\end{tcolorbox}



\begin{tcolorbox}[width=\textwidth, breakable]
{
\renewcommand{\lstlistingname}{Prompt}
\begin{lstlisting}[
    frame=none,
    basicstyle=\ttfamily\footnotesize,
    breaklines=true,
    breakatwhitespace=true,
    breakindent=0pt,
    columns=fullflexible,
    caption={Example of Script-Style Intent Clustering Knowledge (TIC)},
    captionpos=b,
    label={prompt:TIC}
]
Original Question:
I've just secured all doors and engaged the parking brake in my vehicle, please start the engine with the ignition on START mode.

Enhanced Question:
I've just secured all doors and engaged the parking brake in my vehicle, please start the engine with the ignition on START mode.

Before answering the user's question above, please first review the following related experiences:

The user's intent is The user has secured the vehicle's doors and engaged the parking brake, and is requesting to start the engine by turning the ignition to the START mode.
There are some behavior pattern for you to reference:
**Pattern Summary of The common intent is to ensure car doors are locked and the engine is started for a safe and prepared journey.**: Across the provided tool call paths, the mainstream pattern is to secure the vehicle by locking all doors, then start the engine by fully pressing the brake pedal and invoking START. Variations include checking door status before locking, engaging the parking brake for safety, turning on headlights when requested, and performing post-start checks like climate or tire pressure. Common pitfalls observed: skipping a pre-check of door status before locking; not verifying the outcome of door lock/unlock; not confirming the engine actually started; neglecting fuel/battery checks if start fails; forgetting to release the brake pedal and/or parking brake before driving; failing to handle low tire pressure; and not confirming or setting headlights appropriately for conditions. To make the workflow robust, it should include: door status verification and confirmation after lock/unlock; parking brake engagement logic (engage when stationary or on slopes, release when ready to drive); brake pedal press verification; engine start verification and diagnostics on failure (fuel/battery checks, possible refuel); optional safety checks (tire pressure); optional lighting/climate steps; and clean-up steps like releasing the brake and parking brake before departure.
**Note**: You are not required to reference the information or examples above if they are not directly relevant to the current user question. Analyze the problem carefully, decide whether the retrieved information is useful, and always apply reasoning before making any tool calls.
Your actions must be based on the information given by the current user. You can not make up data, nor can you refer to examples that will cause you to act beyond the current information.
You need to determine the difference between your question and the question in retrieval examples.
Attention the user question at current turn is: 
I've just secured all doors and engaged the parking brake in my vehicle, please start the engine with the ignition on START mode.
\end{lstlisting}
}
\end{tcolorbox}


\end{document}